\documentclass[journal]{IEEEtai}

\usepackage{inputenc}
\usepackage{scalerel}
 \usepackage{xcolor}

\let\labelindent\relax

\usepackage[margin=1in]{geometry}

\usepackage{url}            
\usepackage{cite}
\usepackage{hyperref}

\definecolor{darkpastelgreen}{rgb}{0.01, 0.75, 0.24}
	\definecolor{cadmiumgreen}{rgb}{0.0, 0.42, 0.24}
\definecolor{armygreen}{rgb}{0.29, 0.33, 0.13}
\hypersetup{colorlinks,linkcolor={blue},citecolor={darkpastelgreen},urlcolor={red}}

\usepackage{booktabs}       % professional-quality tables
\usepackage{amsfonts}       % blackboard math symbols
\usepackage{nicefrac}       % compact symbols for 1/2, etc.
\usepackage{microtype}      % microtypography
\usepackage{url}
\usepackage{array}

\usepackage{amsmath}
\usepackage{color}
\usepackage{amssymb}
\usepackage{amsthm}
\usepackage{pifont}% http://ctan.org/pkg/pifont
\usepackage[linesnumbered,ruled,vlined]{algorithm2e}

\usepackage{paralist}
\usepackage{dsfont}

\usepackage[square,sort,comma,numbers]{natbib}
\usepackage{csquotes}
\usepackage{comment}
\usepackage{mathtools}

\usepackage{microtype}
\usepackage{graphicx}
\usepackage{booktabs} % for professional tables
\usepackage{dirtytalk}
\usepackage{subfiles}

\usepackage{caption}
\usepackage{subcaption}

\usepackage{xspace}
\usepackage{forloop}
\usepackage{multirow}
\usepackage{algorithm,algorithmic,refcount}
\usepackage{comment}

\usepackage{graphicx}
\usepackage[shortlabels]{enumitem}
\usepackage{bbm}
\usepackage{float}
\usepackage[most]{tcolorbox}

\usepackage{footnote}
\usepackage{etoolbox}
\BeforeBeginEnvironment{tcolorbox}{\savenotes}
\AfterEndEnvironment{tcolorbox}{\spewnotes}

\usepackage[mathscr]{euscript}
\usepackage{accents}

\DeclareSymbolFont{rsfs}{U}{rsfs}{m}{n}
\DeclareSymbolFontAlphabet{\mathscrsfs}{rsfs}

\renewcommand{\P}{\mathbb{P}}
\newcommand{\E}{\mathbb{E}}

\def\bW{{\boldsymbol W}}

\def\bx{{\boldsymbol x}}

\def\btheta{{\boldsymbol \theta}}

\usepackage{color}

\newcommand{\argmin}[1]{\underset{#1}{\mathrm{argmin}} \ }

\definecolor{red}{gray}{0}

\begin{document}

%\title{FedZoo-Bench: A Practical Recipe for Federated Learning with Non-IID Data Experimental Design} 

\title{A Practical Recipe for Federated Learning Under Statistical Heterogeneity Experimental Design}

\author{Mahdi Morafah, Weijia Wang, and Bill Lin
\thanks{M. Morafah, W. Wang and B. Lin are all with Electrical and Computer Engineering Department of University of California San Diego, USA (e-mail address: \{mmorafah, wweijia, billlin\}@eng.ucsd.edu).}
\thanks{Corresponding author: Mahdi Morafah.}
\thanks{}
}
%\thanks{This paragraph will include the Associate Editor who handled your paper.}}

\markboth{Journal of IEEE Transactions on Artificial Intelligence, Vol. 00, No. 0, Month 2023}
{M. Morafah \MakeLowercase{\textit{et al.}}: IEEE Journals of IEEE Transactions on Artificial Intelligence}

\maketitle

\begin{abstract}
Federated Learning (FL) has been an area of active research in recent years. There have been numerous studies in FL to make it more successful in \textcolor{red}{the} presence of data heterogeneity. However, despite the existence of many publications, the \textcolor{red}{state} of progress in the field is unknown. Many of the works use inconsistent experimental settings and there \textcolor{red}{are} no comprehensive studies on the effect of FL-specific experimental variables on the results and practical insights for a more comparable and consistent FL experimental setup. Furthermore, \textcolor{red}{the} existence of several benchmarks and confounding variables \textcolor{red}{has} further complicated the issue of inconsistency and ambiguity. In this work, we present the first comprehensive study on the effect of FL-specific experimental variables in relation to each other and performance results, \textcolor{red}{bringing} several insights and recommendations for designing a meaningful and well-incentivized FL experimental setup. We further aid the community by releasing FedZoo-Bench, an open-source library based on PyTorch with pre-implementation of 22 state-of-the-art methods\footnote{We will continue the effort to extend FedZoo-Bench by implementing more methods and adding more features. Any contributions to FedZoo-Bench would be greatly appreciated as well.}, and a broad set of standardized and customizable features available at~\url{https://github.com/MMorafah/FedZoo-Bench}. We also provide a comprehensive comparison of several state-of-the-art (SOTA) methods to better understand the current state of the field and existing limitations.
\end{abstract}

\begin{IEEEImpStatement}
Federated Learning aims to train a machine learning model using the massive decentralized data available at IoT and mobile devices, and different data centers while maintaining data privacy. However, despite {the} existence of numerous works, the state of progress in the field is not well-understood. Papers use different methodologies and experimental setups that \textcolor{red}{are} hard to compare and examine the effectiveness of methods in more general settings. Moreover, the effect of federated learning experimental design factors such as local epochs, and sample rate on the performance results have remained unstudied in the field. Our work comprehensively studies the effect of experimental design factors in federated learning, provides suggestions and insights, introduces FedZoo-Bench with \textcolor{red}{the} pre-implementation of 22 state-of-the-art algorithms under a unified setting, and finally measures the state of progress in the field. The studies and findings discussed in our work can significantly help the federated learning field by providing a more comprehensive understanding of the impact of experimental design factors, facilitating the design of better performing algorithms, and enabling \textcolor{red}{a} more accurate evaluation of the effectiveness of different methods. 
\end{IEEEImpStatement}


%\newpage

%\input{defs.tex}

\section{Introduction}

\IEEEPARstart{F}{ederated Learning} (FL) is a machine learning setting, which aims to collaboratively train machine learning models with the participation of several clients under the orchestration of a central server in a privacy-preserving and communication efficient manner~\cite{mcmahan2017communication, kairouz2021advances}. FL has seen a surge of interest in the machine learning research community in recent years, thanks to its potential to improve the performance of edge users without compromising data privacy. This innovative approach has been successfully applied to a wide range of tasks, including image classification, natural language processing, and more~\cite{kairouz2021advances, yang2018applied, sui2020feded, muhammad2020fedfast, dayan2021federated}.

The ultimate goal of standard (global) FL is to train a shared global model which uniformly performs well over almost the entire participating clients. However, the inherent diversity and not independent and identical (Non-IID) distribution of clients' local data \textcolor{red}{has} made the global FL approach very challenging~\cite{haddadpour2019convergence, li2019feddane, zhao2018federated, hsieh2020non, li2020federated, sattler2020clustered}. Indeed, clients' incentives to participate in FL can be to derive personalized models rather than learning a shared global model. This client-centric perspective along with the challenges in the global FL approach under Non-IID data distributions has motivated an alternative personalized FL approach. Personalized FL aims to learn personalized models performing well according to the distinct data distribution of each participating client.

Despite the significant \textcolor{red}{number} of works that have been done in both global and personalized FL approaches under data heterogeneity, the state of progress is not well understood in the FL community. In particular, the following key questions have remained unanswered in the existing literature: what factors affect experimental results and how to control them? Which experimental design setups are effective and well-incentivized for a fair comparison in each FL approach? What are the best practices and remedies to compare different methods and avoid evaluation failure? We primarily find that the methodologies, experimental setups, and evaluation metrics are so inconsistent between papers that a comprehensive comparison is impossible. For example, some papers consider a specific FL approach, however, they use an experimental setup \textcolor{red}{that} is not well-incentivized or has been created to match their assumptions and \textcolor{red}{is} not applicable to other cases. Moreover, the existence of numerous benchmarks and inconsistent implementation environments together with different confounding variables such as data augmentation and pre-processing techniques, choice of \textcolor{red}{the} optimizer, and learning rate schedule have made such comparison even more difficult.

To address the mentioned issues in the current state of FL research, we present the first comprehensive study, to the best of our knowledge, on FL-specific experimental variables and provide new insights and best practices for a meaningful and well-incentivized FL experimental design for each FL approach. We also introduce FedZoo-Bench, an open-source library based on PyTorch that provides a commonly used set of standardized and customizable features in FL, and implementation of 22 state-of-the-art (SOTA) methods under a unified setting to make FL research more reproducible and comparable. Finally, we present a comprehensive evaluation of several SOTA methods in terms of performance, fairness, and generalization to newcomers, to provide a clear understanding of the promises and limitations of existing methods.

\textbf{Contributions.} Our study makes the following key contributions:
\begin{itemize}
\item We conduct the first comprehensive analysis of FL experimental design variables by running extensive experiments and provide new insights and best practices for a meaningful and well-incentivized experimental setup for each FL approach.
\item We introduce FedZoo-Bench, an open-source library for implementation and evaluation of FL algorithms under data heterogeneity, \textcolor{red}{consisting of the} implementation of 22 SOTA algorithms, available at~\url{https://github.com/MMorafah/FedZoo-Bench}.
\item Using FedZoo-Bench, we conduct a comprehensive evaluation of several SOTA algorithms in terms of performance, fairness, and generalization to newcomers to form a clear understanding of the strengths and limitations of existing methods.
\end{itemize}

% \textbf{Contributions.} We provide a summery of our contributions bellow:
% \begin{itemize}
%     \item We provide the first comprehensive study to the best of our knowledge on FL-specific experimental design variables by running more than XX different experiments.
%     \item We identify insights and recommendations for a more meaningful and well-incentivized experimental setup for each FL approach.
%     \item We develop FedZoo-Bench, an open-source library for implementation and evaluation of FL algorithms under data heterogeneity containing implementations of more than XX different algorithms available at XX.
%     \item We utilize the best practices that we identify and experimentally evaluate 17 different state-of-the-art (SOTA) algorithms under a common configurations in terms of performance, fairness and generalization to newcomers to better understand the state of progress for each FL approach.
    
%     %\item Identify and agrregate different reported methodologies and protocols in federated learning with Non-IID data literature based on our study of 200 papers. 
%     %\item We provide a list of experimental inconsistencies in the literature, insights and best protocols to avoid them.
% \end{itemize}

\textbf{Organization.} The rest of the paper is organized as follows. In Section~\ref{related-works} we bring a concise literature review. In Section~\ref{background} we provide the background for each FL approach and statistical heterogeneity. In Section~\ref{study} we provide our comprehensive study on FL-specific experimental variables. In Section~\ref{recommendation} we \textcolor{red}{discuss our recommendations.} In Section~\ref{fedzoo} we introduce FedZoo-Bench and compare 17 different algorithms. In Section~\ref{sec:conclusion} we conclude and provide future works.
\section{Literature Review}\label{related-works}
%There has been a growing interest in federated learning recently and the number of works are proliferating. In this section we provide a concise review of the FL literature.

\textbf{Global FL.} McMahan et al.~\cite{mcmahan2017communication} proposed the first method for global FL called FedAvg which simply averages the local models to update the server-side model. This FL approach mainly suffers from poor convergence and degradation of the results in \textcolor{red}{the} presence of data heterogeneity~\cite{tan2022towards, zhao2018federated}. Some works attempt to address these issues by regularizing local training. FedProx~\cite{li2020federated} uses an L2 regularizer on the weight difference between the local and global model. MOON~\cite{li2021model} utilizes contrastive learning to preserve global knowledge during local training. In FedDyn~\cite{acar2021federated}, an on\textcolor{red}{-}device dynamic regularization at each round has been used to align the local and global solutions. Another set of works \textcolor{red}{studies} the optimization issues of FedAvg in the presence of data heterogeneity and \textcolor{red}{proposes} alternative optimization procedures with better convergence guarantees~\cite{wang2020tackling, karimireddy2020scaffold, karimireddy2020mime, reddi2020adaptive, haddadpour2019convergence}. FedNova~\cite{wang2020tackling} addresses the objective inconsistency caused by heterogeneity in the local updates by weighting the local models in server-side averaging to eliminate bias in the solution and achieve fast error convergence. Scaffold~\cite{karimireddy2020scaffold}  proposes control variates to correct the local updates and eliminate the ``client drift'' which happens because of data heterogeneity resulting \textcolor{red}{in} convergence rate improvement. Other approaches have focused on proposing better model fusion techniques to improve performance~\cite{lin2020ensemble, wang2020federated, liu2022deep, sattler2021fedaux, gong2021ensemble}. FedDF~\cite{lin2020ensemble} adds a server-side KL training step after averaging local models by using the average of clients' logits on a public dataset. FedMA~\cite{wang2020federated} proposes averaging on the matched neurons of the models at each layer. GAMF~\cite{liu2022deep} formulates model fusion as a graph matching task by considering neurons or channels as nodes and weights as edges. For a more detailed review of methods on global FL literature, we recommend reading the surveys~\cite{mahlool2022comprehensive, kairouz2019advances}.

% There are some studies through bayesian learnig~\cite{}, hyper-networks~\cite{}, batch-normalization~\cite{} as well.

\textbf{Personalized FL.} The fundamental challenges of \textcolor{red}{the} global FL approach, such as poor convergence in the presence of data heterogeneity and lack of personalized solutions, have motivated the development of personalized FL. Personalized FL aims to obtain personalized models for participating clients. There are various efforts to solve this problem through different techniques. Multi-Task Learning (MTL) based techniques have been proposed in a number of works by considering clients as tasks and framing the personalized FL as an MTL problem~\cite{smith2017federated, t2020personalized, hanzely2020lower, li2021ditto, marfoq2021federated, dinh2021fedu}. Another group of studies \textcolor{red}{proposes} model interpolation techniques by mixing the global and local models~\cite{mansour2020three, deng2020adaptive, marfoq2022personalized}. There are also works that utilize representation learning techniques by decoupling parameters (or layers) into global and local parameters and then averaging only the global parameters with other clients~\cite{liang2020think, arivazhagan2019federated, collins2021exploiting, pillutla2022federated}. Additionally, there are some meta-learning\textcolor{red}{-}based works that attempt to obtain a global model with the capability of getting personalized fast by further local fine-tuning~\cite{jiang2019improving, fallah2020personalized, fallah2020personalized, chen2018federated, singhal2021federated}. Clustering-based methods have been also shown to be effective in several studies by grouping clients with similar data distribution to achieve better personalized models and faster convergence~\cite{ghosh2020efficient, briggs2020federated, lyu2022personalized, duan2021flexible, cho2021personalized, sattler2020clustered, vahidian2022efficient}. More recently, personalized FL has been realized with pruning-based techniques as well~\cite{li2020lotteryfl, vahidian2021personalized, li2021fedmask, bibikar2022federated, huang2022achieving}. For a more detailed review of methods on personalized FL literature, we recommend the surveys~\cite{kulkarni2020survey, zhu2021federated}.

%FedPer~\cite{arivazhagan2019federated} is very similar to LG-FedAvg with the difference that let clients to have different initializations for the local personalized layers.
%Clustering
%Fine-Tuning/Transfer Learning
%Pruning
%\textbf{FL Benchmarks.} Recently, there has been a growing interest in using federated learning for various applications on different platforms. To support this, several FL benchmarks and codebases have been developed, including FATE~\cite{liu2021fate}, FedML~\cite{he2020fedml}, PaddleFL~\cite{paddlefl}, FedLearner~\cite{fedlearner}, TFF~\cite{tff}, FLOWER~\cite{beutel2020flower}, FLUTE~\cite{dimitriadis2022flute}, FedTree~\cite{li2022fedtree}, Motley~\cite{wu2022motley}, PFL~\cite{chen2022pfl}, and NIID-Bench~\cite{li2021federated}. Many of available benchmarks support only global FL with limited customization for research.  

\textbf{FL Benchmarks.} Current FL benchmarks primarily focus on enabling various platforms and APIs to perform FL for different applications. They often only realize the global FL approach and implement basic algorithms (e.g. FedAvg)~\cite{liu2021fate, paddlefl, tff, fedlearner}, while other benchmarks,such as FLOWER~\cite{beutel2020flower}, FedML~\cite{he2020fedml}, FLUTE~\cite{dimitriadis2022flute} and NIID-Bench~\cite{li2021federated}, offer more customizable features, and implementation of more algorithms. For a more detailed review \textcolor{red}{of} the applicability and comparison \textcolor{red}{of} the existing benchmarks, we defer the reader to UniFed~\cite{liu2022unifed}.  While the majority of existing benchmarks are for global FL, there are \textcolor{red}{a} few recently released personalized FL benchmarks, including pFL-Bench~\cite{chen2022pfl} and Motley~\cite{wu2022motley}. However, these benchmarks do not investigate the effect of experimental variables and only consider a few algorithms and their variants for comparison. In contrast, FedZoo-Bench offers support for both global and personalized FL approaches, providing implementation of 22 SOTA methods and the ability to assess generalization to newcomers. Additionally, our study on the effect of FL-specific experimental variables in relation to each other and performance results, together with \textcolor{red}{the} identification of more meaningful setups for each FL approach, is a new contribution to the field.
\section{Background: Federated Learning and Statistical Heterogeneity} \label{background}
%In this section, we provide the backgrounds for each FL approach and statistical heterogeneity.
\subsection{Overview of Federated Learning and Notations}
Consider a server and a set of clients $S$ which participate in the federation process. $f_i(\bx; \btheta_i)$ is the $L$ layers neural network model of client $i$ with parameters $\btheta_i=(\bW_i^1,\ldots,\bW_i^L)$ where $\bW^l$ stands for the $l$-th layer weights, training dataset $D_i^{train}$, and test dataset $D_i^{test}$. \textcolor{red}{At} the beginning of communication round $t$, a subset of clients $S_t$ are randomly selected with the sampling rate of $C \in (0, 1]$ out of total $N$ available clients. The selected clients receive the parameters from the server and perform local training for $E$ epochs with \textcolor{red}{the} batch size of $B$ and learning rate of $\eta$. At the end of communication round $t$, the selected clients send back their updated parameters to the server for server-side processing and model fusion. This federation process continues for $T$ total communication rounds. Algorithm~\ref{alg:FederatedLearning} shows this process in detail.

\subsection{Problem Formulation}
%We formally define the problem of global and personalized FL in this section. 

\textbf{Global Federated Learning (gFL).}
The objective is to train a shared global model at the server which uniformly performs well on each client and the problem is defined as follow: 
\begin{align} \label{eq:gfl}
    \widehat \btheta_g = \argmin {\btheta_g} \sum_i^N \E_{(\bx,y)\sim D_i^{\text{train}}}[\ell(f_i(\bx; \btheta_g), y)].
\end{align}
FedAvg~\cite{mcmahan2017communication} is the first and most popular algorithm proposed to solve \textcolor{red}{Equation}~\ref{eq:gfl}, which uses parameter averaging at the server-side for model fusion.

\textbf{Personalized Federated Learning (pFL).}
The objective is to train personalized models \textcolor{red}{to perform} well on each client's distinctive data distribution and the problem is defined as \textcolor{red}{follows}: 
\begin{align} \label{eq:pfl}
\textcolor{red}{
    \{{\widehat \btheta_i}\}_1^N = \argmin {\{\btheta_i\}_1^N} \sum_i^N
    \E_{(\bx,y)\sim D_i^{\text{test}}}[\ell(f_i(\bx; \btheta_i), y)].
    }
    %\{\btheta_1,\ldots,\btheta_N\}
\end{align}
FedAvg + Fine-Tunining (FT)~\cite{jiang2019improving} is the simplest algorithm proposed to solve \textcolor{red}{Equation}~\ref{eq:pfl}, where each client fine-tunes the global model obtained via FedAvg on their local data.

\begin{algorithm}[t]
\caption{Federated Learning}
\label{alg:FederatedLearning}

\begin{algorithmic}[1]
\REQUIRE number of clients ($N$), sampling rate ($C\in(0,1]$), number of communication rounds ($T$), local dataset of client $k$ ($D_k$), local epoch ($E$), local batch size ($B$), learning rate ($\eta$).
%\STATE \textbf{Server Executes:}
\STATE Initialize the server model with $\btheta_g^0$ \;
\FOR {each round $t = 0, 1, \ldots, T-1$}
\STATE $m \leftarrow {\rm{max}}(C \cdot N,1)$ \;
\STATE $S_t \leftarrow $(random set of m clients) \; %\tcp*{set of $m$ available clients}
\FOR {each client $k \in S_t$ \rm{\textbf{in parallel}}}
\STATE $\btheta^{t+1}_{k}\leftarrow {\mathtt{ClientUpdate}}(k; \btheta^t_{g} )$ \; 
\ENDFOR
\STATE $\btheta^{t+1}_{g}=\mathtt{ModelFusion}(\{\btheta_k \}_{k \in S_t})$ \COMMENT{FedAvg~\cite{mcmahan2017communication}: $\theta^{t+1}_{g}=\sum_{k \in S_t }{|D_{k}|\theta^{t+1}_{k}}  /\sum_{k \in S_t }{|D_{k}|}$}
%\STATE $\theta^{t+1}_{g}=\sum_{k \in S_t }{|D_{k}|\theta^{t+1}_{k}}  /\sum_{k \in S_t }{|D_{k}|}$
\ENDFOR
\vspace{1mm}
\FUNCTION {$\mathtt{ClientUpdate}(k, \btheta^t_{g})$}
\STATE $\btheta^{t}_k \leftarrow \btheta^t_{g}$ \;
\STATE $\mathcal{B} \leftarrow$ ({randomly splitting $D_k^{train}$ into batches of size $B$})\;
\FOR {each local epoch $ \in \{1, \ldots, E\}$}
\FOR {each batch $\mathbf{b} \in \mathcal{B}$}
\STATE $\btheta^{t}_k \leftarrow \btheta^{t}_k - \eta \nabla_{\btheta^{t}_k} \ell(f_k(\bx; \btheta^{t}_k), y)$\;
\ENDFOR
\ENDFOR
\STATE $\btheta^{t+1}_k \leftarrow \btheta^{t}_k$\;
\ENDFUNCTION
% \Fn{\FClient{$k, \btheta^t_{g}$}}{
%     \STATE $\btheta^{t}_k \leftarrow \btheta^t_{g}$ \;
%     \STATE $\mathcal{B} \leftarrow$ ({randomly splitting $D_k$ into batches of Size $B$})\;
%     \For {each local epoch $i = 1, \dots, E$} {
%         \For {each batch $\mathbf{b} \in \mathcal{B}$} {
%             \STATE $\btheta^{t}_k \leftarrow \btheta^{t}_k - \eta \nabla_{\btheta^{t}_k} f_k(\btheta^{t}_k; \mathbf{b})$\;
%         }
%     }
%     \STATE $\btheta^{t+1}_k \leftarrow \btheta^{t}_k$\;
%     \KwRet $\btheta^{t+1}_{k}$\;
% }
\end{algorithmic}
\end{algorithm}

\subsection{Statistical Heterogeneity}
Consider the local data distribution of clients, denoted as $\P_i(\bx, y)$. Clients can have statistical heterogeneity or Non-IID data distribution w.r.t each other when $\P_i(\bx, y) \neq \P_j(\bx, y)$. The most popular way to realize this heterogeneity in FL is label distribution skew in which clients have different label \textcolor{red}{distributions} of a dataset (i.e. $\P_i(y) \neq \P_j(y)$)\footnote{Other mechanisms to realize statistical heterogeneity do exist, but are less commonly used in the FL community.}. Two commonly used mechanisms to create label distribution skew in FL are:
\begin{itemize}
    \item \textbf{Label skew($p$)~\cite{li2021federated}:} each client receives $p\%$ of the total classes of a dataset at random. The data points of each class are uniformly partitioned amongst the clients owning that class. %\vspace{-0.5cm}
    \item \textbf{Label Dir($\alpha$)~\cite{hsu2019measuring}:} each client draws a random vector with the length of total classes of a dataset from Dirichlet distribution with a concentration parameter $\alpha$. The data points of each class are uniformly partitioned amongst clients according to each client's class proportions.
\end{itemize}
\section{Comprehensive Study on FL Experimental Variables}\label{study}
%\subsection{Overview and Setup}
\begin{comment}
In the previous sections, we discussed that there is a lack of consistent and experimental design settings in the literature and prior works tend to create the settings in a way to satisfy their assumptions and show better results. In summary, prior works tend to 
\begin{itemize}
    \item make it unclear about the exact experimental settings, and evaluation metrics
    \item do not follow a exact experimental settings universally
    \item not consider the effect of confounding variables in their results.
\end{itemize}

These obstacles have made it difficult for a fair and comprehensive evaluation of the proposed methods and identify the-state-of-the-art baselines. To identify the best experimental design settings and suggestions, we perform a comprehensive study on different experimental design factors and bring our findings in this section.
\end{comment} 
%In this section, we conduct a comprehensive study on the impact of various experimental variables in FL.
\vspace{-0.2cm}

\textbf{Overview.}  To design an effective FL experiment, it is crucial to understand how the FL-specific variables which are clients' data partitioning (type and level of statistical heterogeneity), local epochs ($E$), sample rate ($C$), and communication rounds ($T$) interact with each other and can affect the results. While communication rounds ($T$) serve as the equivalent of epochs in traditional centralized training and primarily determine the training budget, the other variables have a more direct impact on performance. Hence, we focus our analysis on these variables, in relation to each other and performance results, and evaluation metric failure and derive new insights for the FL community to design meaningful and well-incentivized FL experiments.

%Our analysis of these variables, in relation to each other and performance results, provides new insights and best practices for the FL community to design meaningful and well-incentivized FL experiments and avoid evaluation metric failure.
%We also identify the causes of evaluation metrics failure for each FL approach and bring our suggestions on how to avoid them. The study that we provide in this section derives new insights for the broader FL community. 
%We also bring our findings and suggestions on how to control the effect of each variables for a fair experimental design.

\textbf{Baselines.} We use three key baselines in our study: \emph{FedAvg}~\cite{mcmahan2017communication}, the standard FL baseline that has been widely used in the existing literature and can serve as a good representative for the global FL approach~\footnote{\label{foot1}We show in Section~\ref{sec:comparison} that the performance of this baseline is competitive to the SOTA methods.}; \emph{FedAvg + Fine-Tuning (FT)}~\cite{jiang2019improving}, a simple personalized FL baseline that has been shown to perform well in practice and can serve as a representative for the personalized FL approach~\footref{foot1}; and \emph{SOLO}, a solo (local only) training of each client on their own dataset without participation in federation, which serves as a baseline to evaluate the benefit of federation under different experimental conditions. 
%Our results show that the performance of these baselines is competitive with state-of-the-art methods, indicating that our studies can be generalized to both global and personalized FL approaches.

%We use three important baselines for our study in this section. \emph{FedAvg}~\cite{mcmahan2017communication} which is the standard FL baseline that has been compared with in almost the entire of existing works and can be used as a good representative for the global FL approach~\footnote{\label{foot1}We show in Section~\ref{sec:comparison} that the performance of this baseline is competitive to the SOTA methods and therefore, our studies can be generalized to global and personalized FL approaches.}.
%widely used in the real world applications~\cite{apple-siri}. \emph{FedAvg + Fine-Tuning (FT)}~\cite{jiang2019improving} which is the simplest personalized federated learning baseline and has been shown performing well in practice~\cite{wang2019federated,sim2021robust}. This baseline can be also served as a good representative for the personalized FL approach~\footref{foot1}. \emph{SOLO} which is a simple solo (local only) training of each client on their own dataset without participation in federation and can be competitive when clients have enough data or the statistical heterogeneity level is very high. This serves as a good comparison baseline to identify if federation has enough incentive under a specific experimental setting. 

\textbf{Setup.} We use CIFAR-10~\cite{krizhevsky2009learning} dataset and LeNet-5~\cite{lecun1998gradient} architecture which has been used in the majority of existing works. We fix number of clients ($N $), communication rounds ($R$), and sample rate ($C$) to $100$, $100$, \text{and} $0.1$ respectively; unless specified otherwise. We use SGD optimizer with \textcolor{red}{a} learning rate of $0.01$, and momentum of $0.9$~\footnote{This optimizer and learning rate have been used in some works~\cite{collins2021exploiting, li2021federated, vahidian2022efficient} under a similar setup and we also find that it works the best for our studies.}. We use this base setting for all of our experimental \textcolor{red}{studies} in this section. The reported results are the average results over 3 independent and different runs for a more fair and robust assessment.
% This is general setting which has been used in the majority of existing works as well.

\begin{figure*}[ht]
    \centering
    \begin{subfigure}[b]{0.32\textwidth}
         \centering
         \includegraphics[width=.7\linewidth]{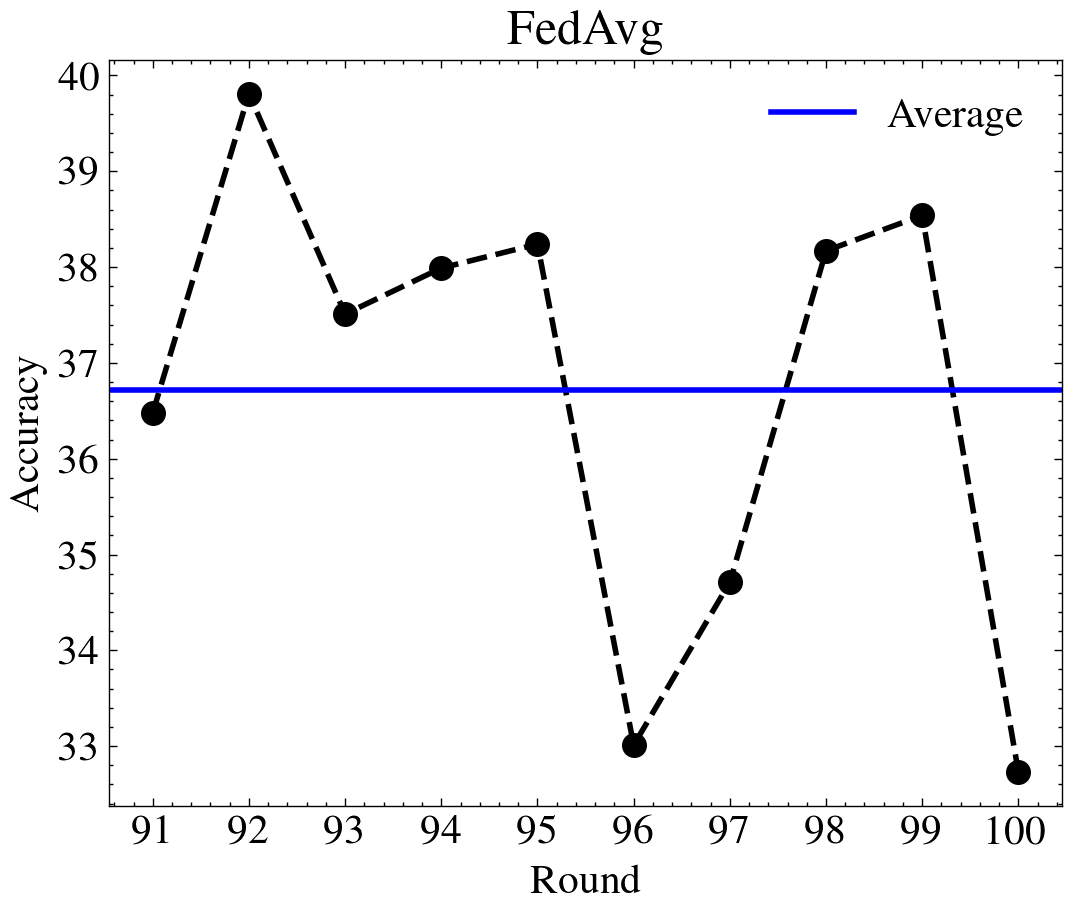}
         \caption{}
         \label{fig1:a}
     \end{subfigure}
     \begin{subfigure}[b]{0.32\textwidth}
         \centering
         \includegraphics[width=.7\linewidth]{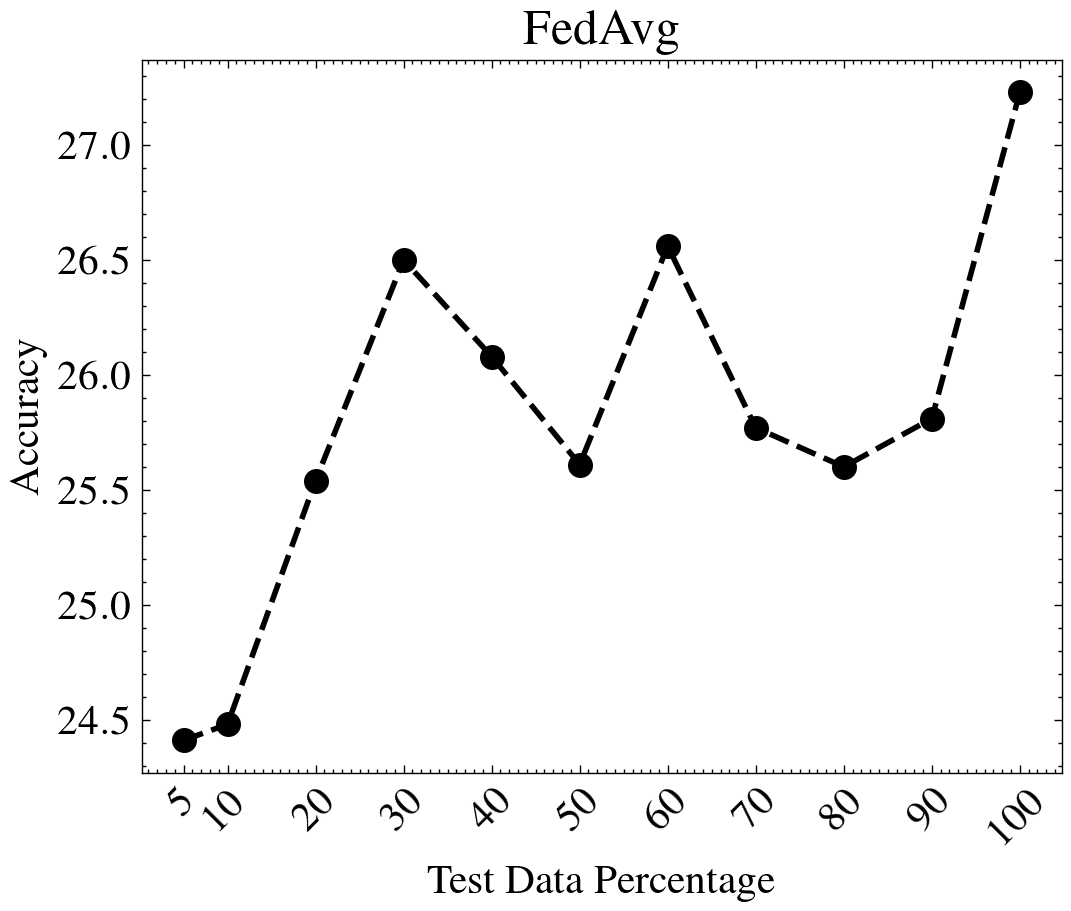}
         \caption{}
         \label{fig1:b}
     \end{subfigure}
      \begin{subfigure}[b]{0.32\textwidth}
         \centering
         \includegraphics[width=.7\linewidth]{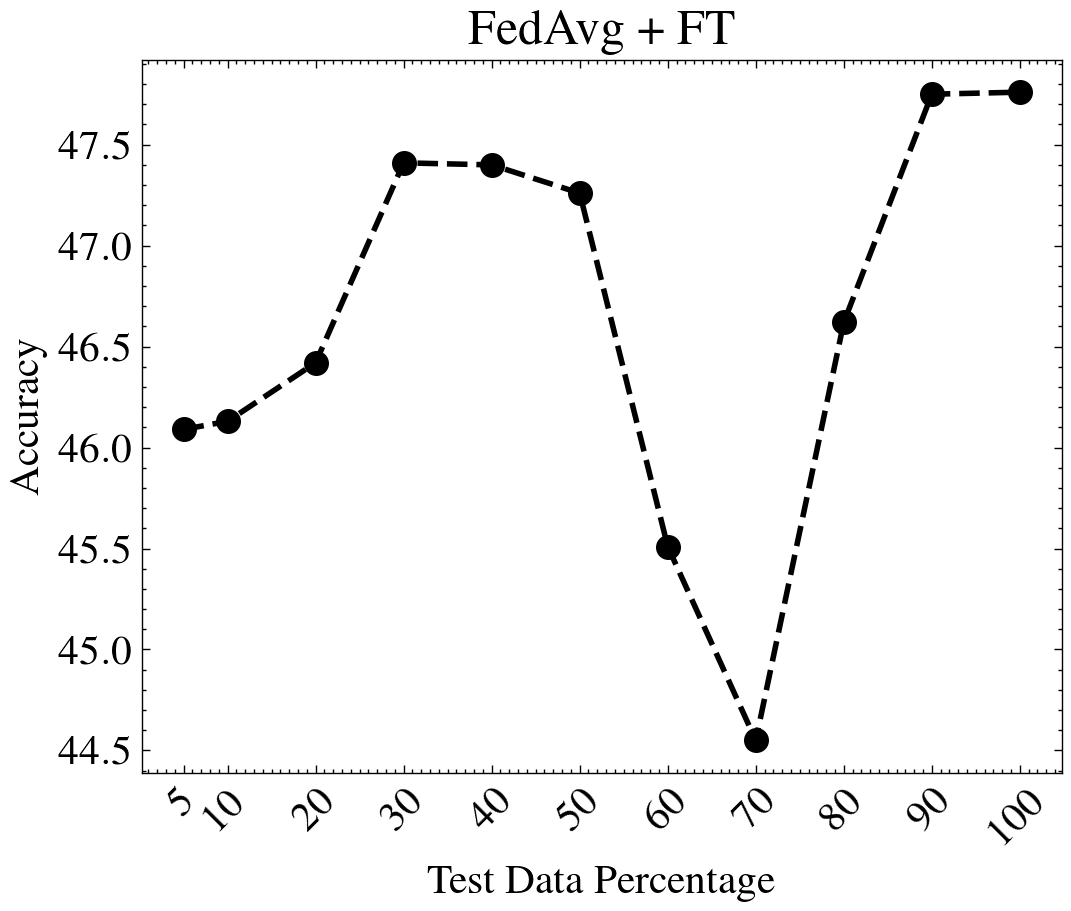}
         \caption{}
         \label{fig1:c}
     \end{subfigure}

    \caption{These figures show factors causing evaluation failures when reporting results: (a) \textcolor{red}{the} accuracy variation of FedAvg over the last $10$ communication rounds, (b)  the variation of FedAvg's accuracy using different test data percentages, (c)  the variation of FedAvg + FT's accuracy using different local test data percentages.}
    \label{fig1}
    \vspace{-0.4cm}
\end{figure*}

\subsection{Evaluation Metric}
\vspace{-0.1cm}
\textcolor{red}{The evaluation metric} for performance is a critical factor in making a fair and consistent assessment in FL. However, the way in which evaluation metrics are calculated in the current FL literature is often ambiguous and varies significantly between papers. In this part we focus on identifying the causes of evaluation metric failures for each FL approach and bring our suggestions for avoiding them.

\textbf{Global FL.} The evaluation metric is the performance of \textcolor{red}{the} global model on the test dataset at the server. We find that the causes for evaluation failures are (1) the round used to report the result and (2) the test data percentage used to evaluate the model. Figure~\ref{fig1:a} shows the global model accuracy over the last $10$ rounds on Non-IID Label Dir($0.1$) partitioning. We can see there is a maximum variation of $7\%$ in the results based on which round to pick for reporting the result. Also, the difference of \textcolor{red}{the} final round result with the average bar is about $4\%$. This shows that the round used to report the result is important and to have a more robust metric \textcolor{red}{for} these variations, it is better to report the average results over a number of rounds. Figure~\ref{fig1:b} shows the variations of the reported result for the same model using different test data percentages. This clearly shows that using different test data points can cause bias in the evaluation. To avoid the mentioned failures and have a more reliable evaluation metric we suggest the following definition:

\begin{tcolorbox}[colback=white!5!white,colframe=black!75!black]\label{def:gfl}
  \textbf{Definition 1} (global FL evaluation metric). We define the average performance of \textcolor{red}{the} global model on the entire test data at the server (if available) over the last $[C \cdot N]$ communication rounds as the evaluation metric for global FL approach~\footnote{\label{foot2}We use this metric definition for all of our experiments.}.
\end{tcolorbox}
\vspace{-0.2cm}
\textbf{Personalized FL.} The evaluation metric is the final average performance of all participating clients on their local test data. The factor which can cause evaluation failure is the local test data percentage used to evaluate each client's model. \textcolor{red}{Prior works have allocated different amounts of data as local test sets to individual clients.} Figure~\ref{fig1:c} shows the variability of the reported results for the same clients under different local test data percentages on Non-IID Label Dir($0.1$) partitioning. This highlights that the use of a randomly selected portion for test data can lead to inaccurate and biased evaluations based on the selected data points be easy or hard.
%\footnote{Previous works, use different percentage of test data for each client when partitioning the dataset across clients.} 
To avoid the mentioned failure in the evaluation metric we suggest the following definition:
\vspace{-0.1cm}
\begin{tcolorbox}[colback=white!5!white,colframe=black!75!black]\label{def:pfl}
  \textbf{Definition 2} (personalized FL evaluation metric). We define the average of \textcolor{red}{the} final performance of all the clients on their entire local test data (if available) as the evaluation metric for personalized FL approach~\footref{foot2}\textsuperscript{,}\footnote{\textcolor{red}{The entire local test data consists of allocating all available test samples belonging to the classes owned by the client.}}.
\end{tcolorbox}

\subsection{Statistical Heterogeneity and Local Epochs} \label{sec:heterogeneity-localepoch}
In this part, we focus our study to understand how different \textcolor{red}{levels and types} of statistical heterogeneity together with local epochs can affect the results and change the globalization and personalization incentives in FL.

\begin{figure*}[ht]
    \centering
    \begin{subfigure}{.32\textwidth}
         \centering
         \includegraphics[width=.7\linewidth]{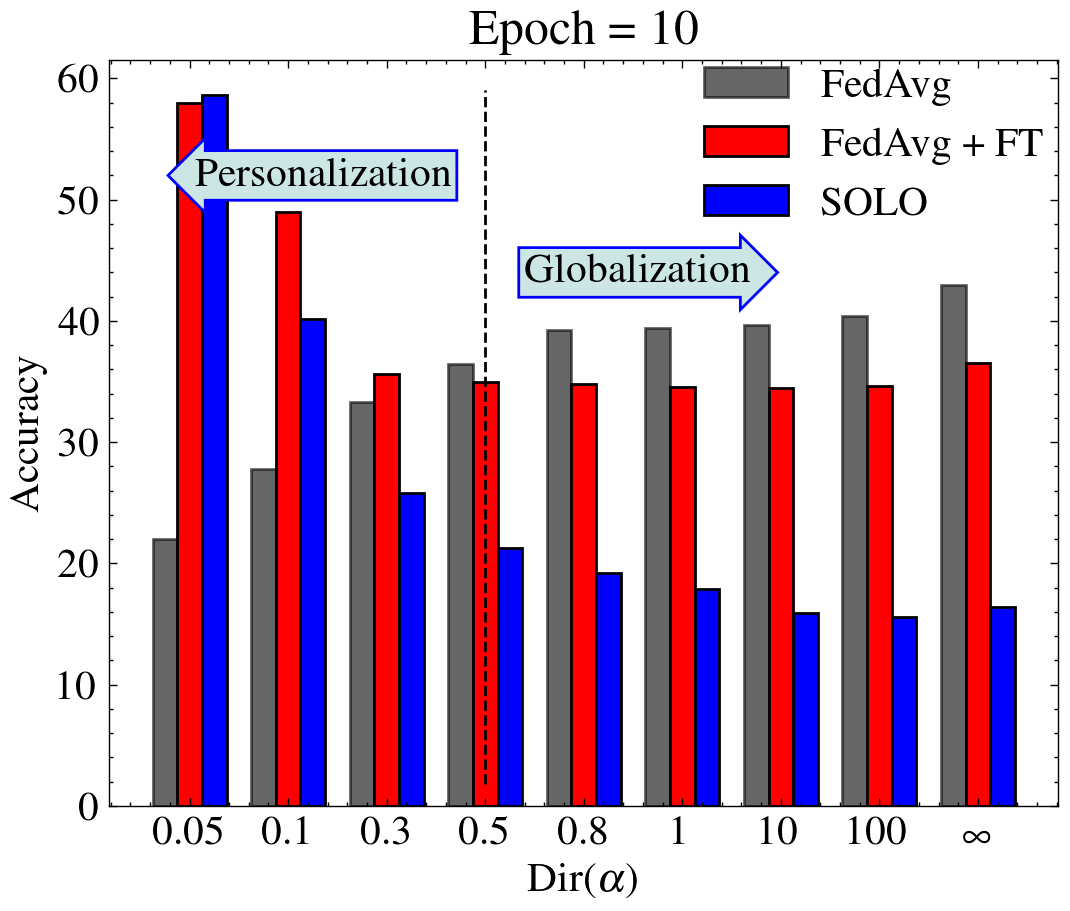}
         \caption{}
         \label{fig2:a}
     \end{subfigure}
     \begin{subfigure}{.32\textwidth}
         \centering
         \includegraphics[width=.7\linewidth]{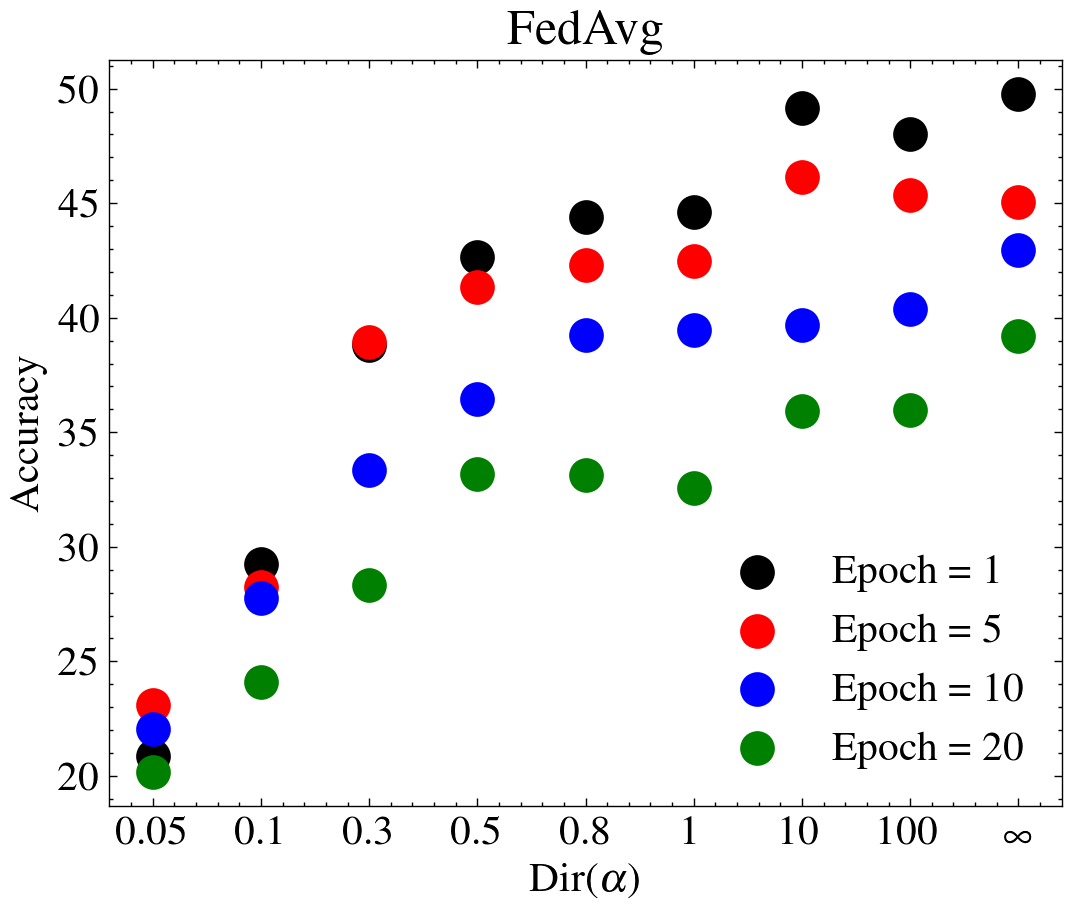}
         \caption{}
         \label{fig2:b}
     \end{subfigure}
     \begin{subfigure}{.32\textwidth}
         \centering
         \includegraphics[width=.7\linewidth]{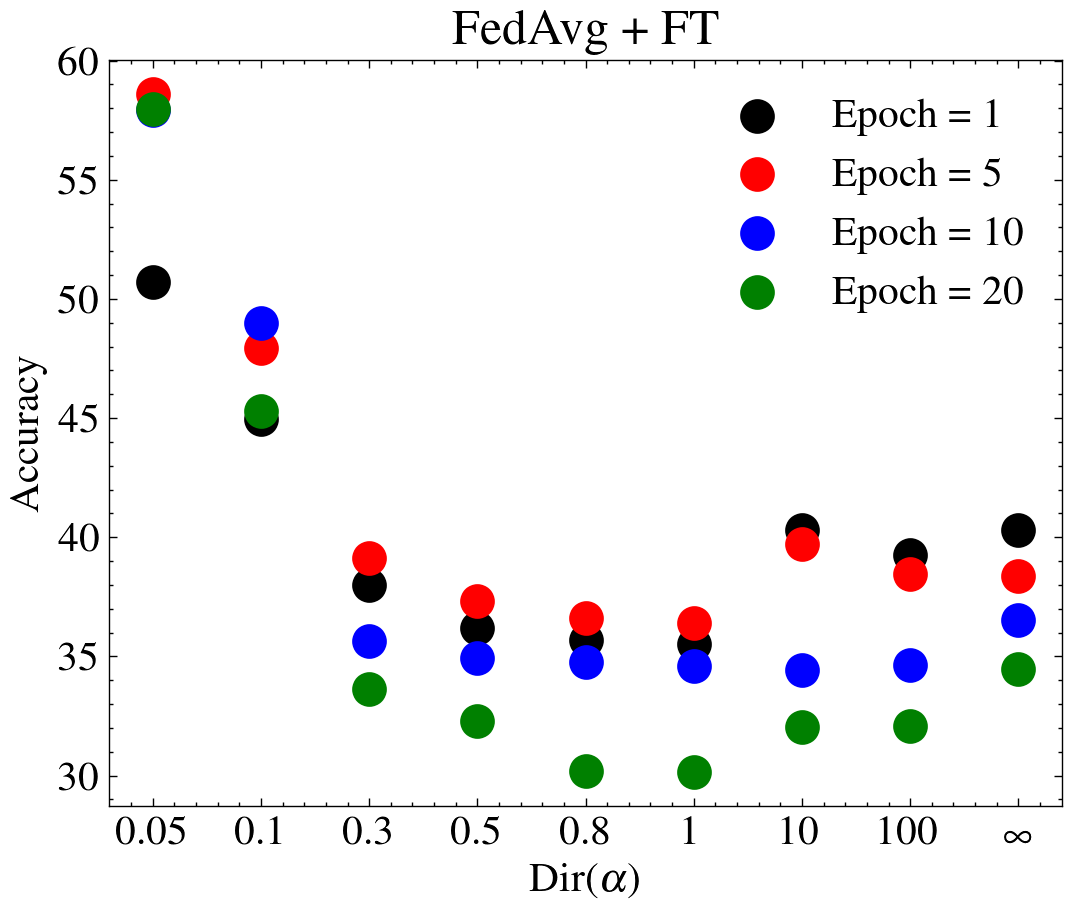}
         \caption{}
         \label{fig2:c}
     \end{subfigure}
    \vspace{-0.3cm}

    \caption{These figures show \textcolor{red}{the effect of the level} of statistical heterogeneity and the number of local epochs for the label Dir type of statistical heterogeneity: (a) illustrates the incentives for globalization and personalization under various levels of statistical heterogeneity with a fixed number of $10$ local epochs, (b) and (c) show the performance of FedAvg and FedAvg + FT under different levels of statistical heterogeneity and local epochs, respectively.}
    \label{fig2}
\end{figure*}

\textbf{Level of statistical heterogeneity.} Figure~\ref{fig2:a} illustrates how the performance of the baselines for a fixed label Dir type of statistical heterogeneity with 10 local epochs varies with the level of statistical heterogeneity. As the level of statistical heterogeneity decreases, the global FL approach becomes more successful than the personalized one. The vertical line in the plot indicates the approximate boundary between the incentives of the two FL perspectives. We can see that in the extreme Non-IID case ($\alpha=0.05$), neither of the FL approaches \textcolor{red}{is} motivated, as the performance of the SOLO baseline is competitive. Additionally, from $\alpha=0.8$ onwards, the global FL approach seems to perform \textcolor{red}{close} to the end of the spectrum at $\alpha=\infty$, which is IID partitioning. Furthermore, we find that the incentives for globalization and personalization can vary with changes in the number of local epochs for a fixed type of statistical heterogeneity (see Section~\ref{appendix:incentives} for more results).

\textbf{Local epochs.} Figures~\ref{fig2:b}, and~\ref{fig2:c} show how the performance of FedAvg and FedAvg + FT for a fixed label Dir type of statistical heterogeneity varies with different \textcolor{red}{levels} of statistical heterogeneity and \textcolor{red}{the} number of local epochs. Figure~\ref{fig2:b} suggests that FedAvg favors fewer local epochs to achieve higher performance. However, Figure~\ref{fig2:c} suggests that FedAvg + FT favors more local epochs for achieving better results but no more than $5$ or $10$ depending on the level of statistical heterogeneity. Our findings support the observation in~\cite{karimireddy2020scaffold} that client drift can have a significant impact on performance, and increasing the number of local epochs amplifies this effect in the results.
%Generally, for both approaches in FL it is desirable to have more local training while not hurting the results which is an aspect that researchers need to consider in their algorithm design.

%Local epoch of $\{5, 10\}$, and $\{1, 5\}$ seems to work really good for Non-IID Label Dir, and Non-IID Label Skew respectively. Another observation is that excessive fine-tuning more than $10$ local epochs can hurt the results. 
% In FL, it is desirable to have more local training and fewer communication rounds while not hurting the results. Increasing the number of local epochs can magnify the effect of client drift which is an aspect that researchers need to consider to handle in their algorithm desing.

\begin{figure*}[h]
    \centering
    \begin{subfigure}{.32\textwidth}
         \centering
         \includegraphics[width=.7\linewidth]{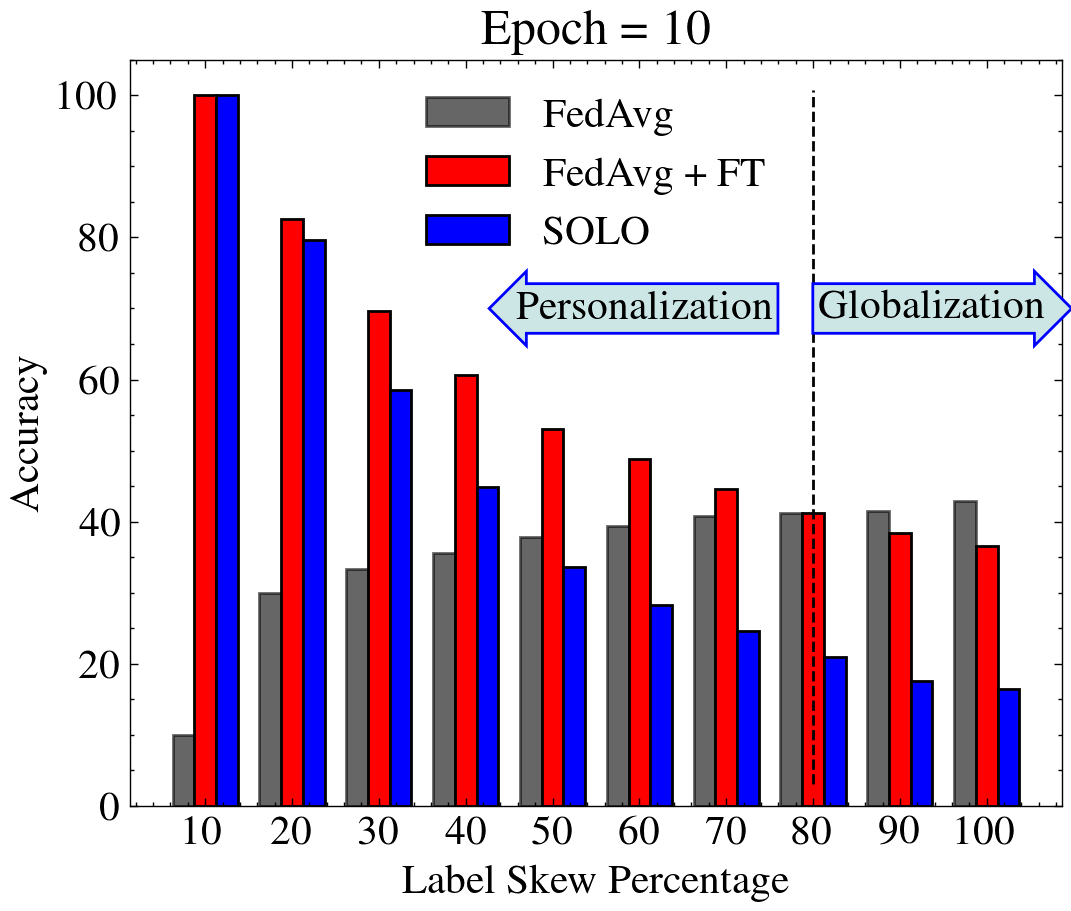}
         \caption{}
         \label{fig3:a}
     \end{subfigure}
     \begin{subfigure}{.32\textwidth}
         \centering
         \includegraphics[width=.7\linewidth]{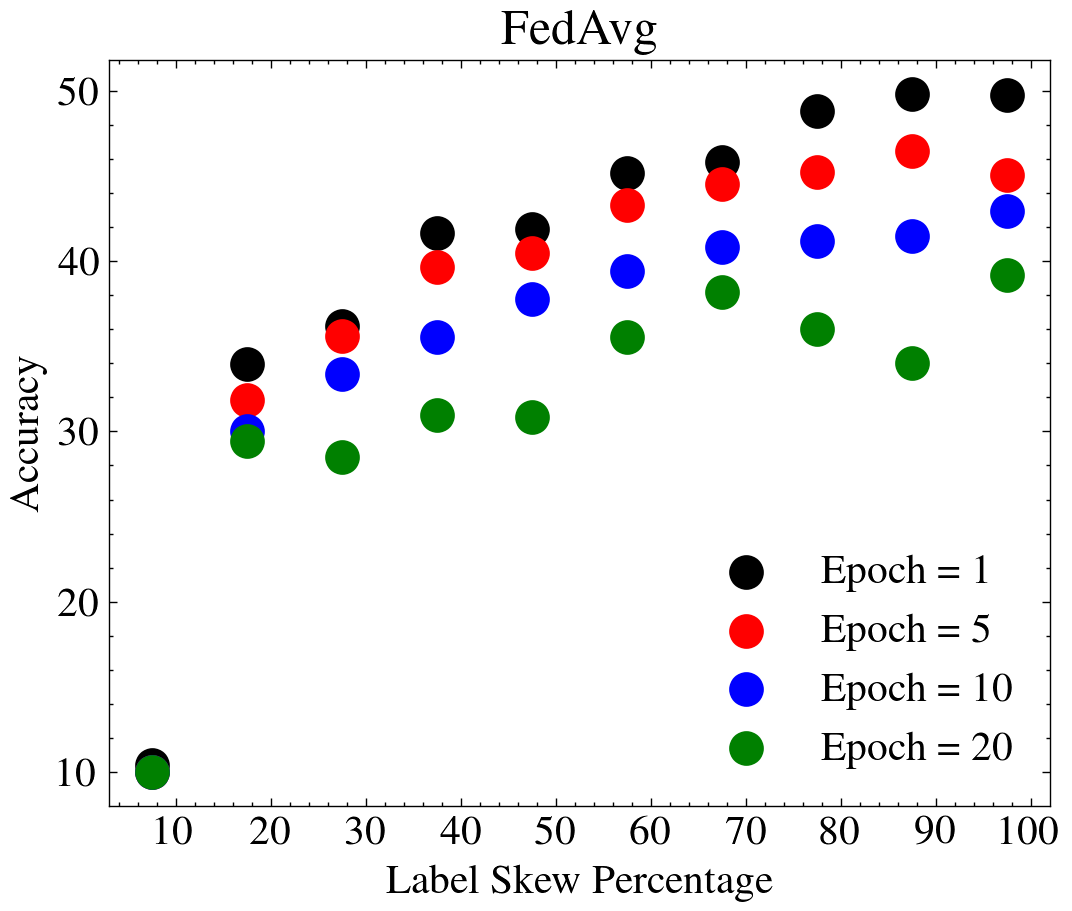}
         \caption{}
         \label{fig3:b}
     \end{subfigure}
     \begin{subfigure}{.32\textwidth}
         \centering
         \includegraphics[width=.7\linewidth]{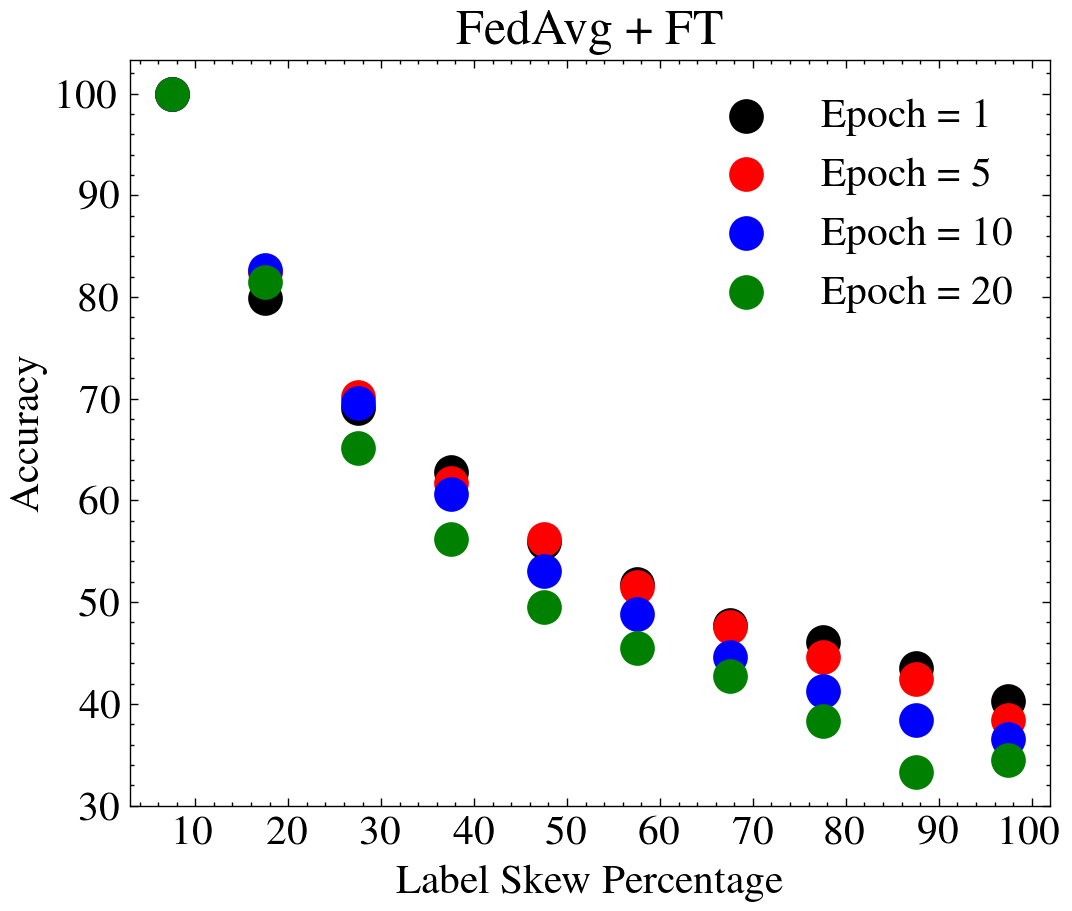}
         \caption{}
         \label{fig3:c}
     \end{subfigure}
    \vspace{-0.3cm}
    \caption{These figures show \textcolor{red}{the effect of the level} of statistical heterogeneity and the number of local epochs for the label skew type of statistical heterogeneity: (a) illustrates the incentives for globalization and personalization under various levels of statistical heterogeneity with a fixed number of $10$ local epochs, (b) and (c) show the performance of FedAvg and FedAvg + FT under different levels of statistical heterogeneity and local epochs, respectively.}
    \label{fig3}
    \vspace{-0.4cm}
\end{figure*}

\textbf{Type of statistical heterogeneity.} Figure~\ref{fig3} illustrates the results of an experiment similar to that of Figure~\ref{fig2}, but with a label skew type of statistical heterogeneity. Figure~\ref{fig3:a} shows the performance of the baselines with fixed $10$ local epochs at different levels of statistical heterogeneity. Comparing this figure with Figure~\ref{fig2:a} reveals that this type of heterogeneity favors personalization over globalization across a wider range of heterogeneity levels. Figures~\ref{fig3:b} and~\ref{fig3:c} show the performance of FedAvg and FedAvg + FT with different levels of statistical heterogeneity and local epochs. Comparing these figures with Figures~\ref{fig2:b} and~\ref{fig2:c} for the label Dir type of statistical heterogeneity reveals that this type of statistical heterogeneity is less affected by an increase in local epochs at each level of heterogeneity. This highlights another finding that the effect of client drift may vary for different types of statistical heterogeneity (see Section~\ref{appendix:incentives} for more results).

\subsection{Sample Rate}
\begin{figure*}[h]
    \centering
    \begin{subfigure}{.24\textwidth}
         \centering
         \includegraphics[width=.85\linewidth]{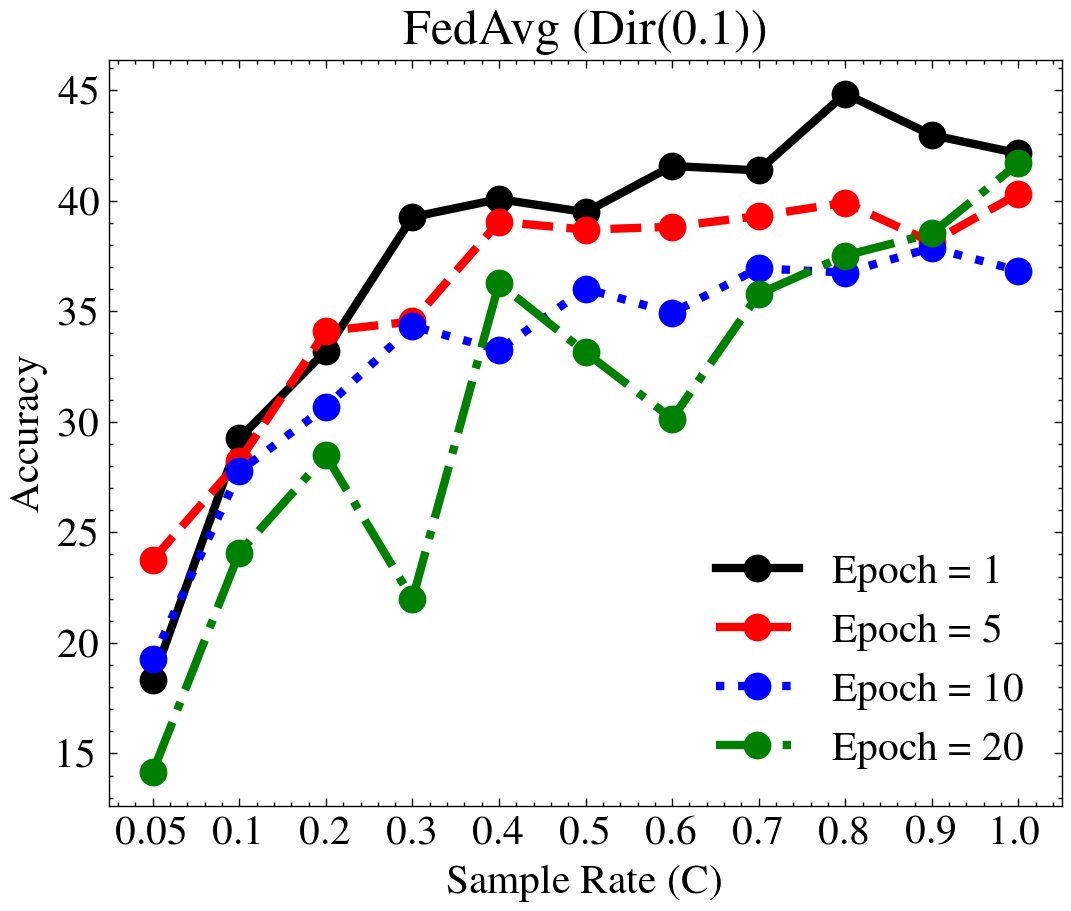}
         \caption{}
         \label{fig5:a}
     \end{subfigure}
     \begin{subfigure}{.24\textwidth}
         \centering
         \includegraphics[width=.85\linewidth]{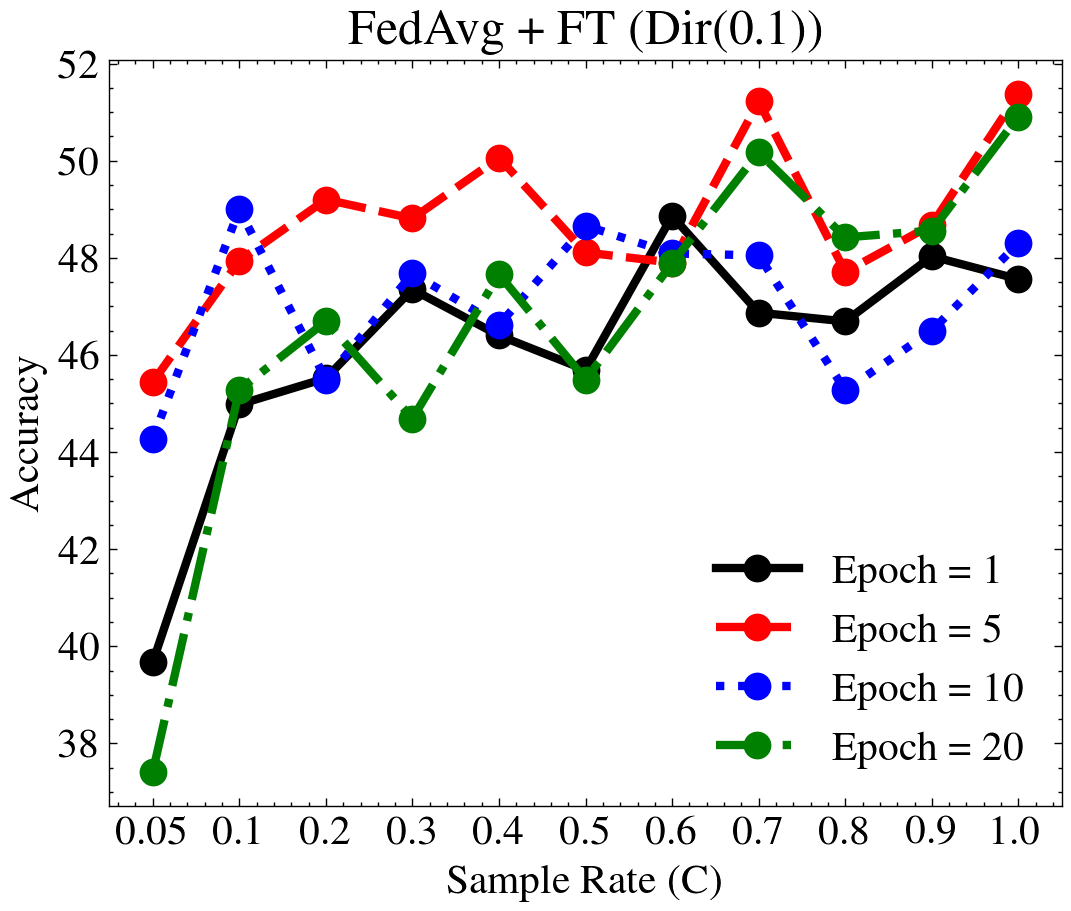}
         \caption{}
         \label{fig5:b}
     \end{subfigure}
     \begin{subfigure}{.24\textwidth}
         \centering
         \includegraphics[width=.85\linewidth]{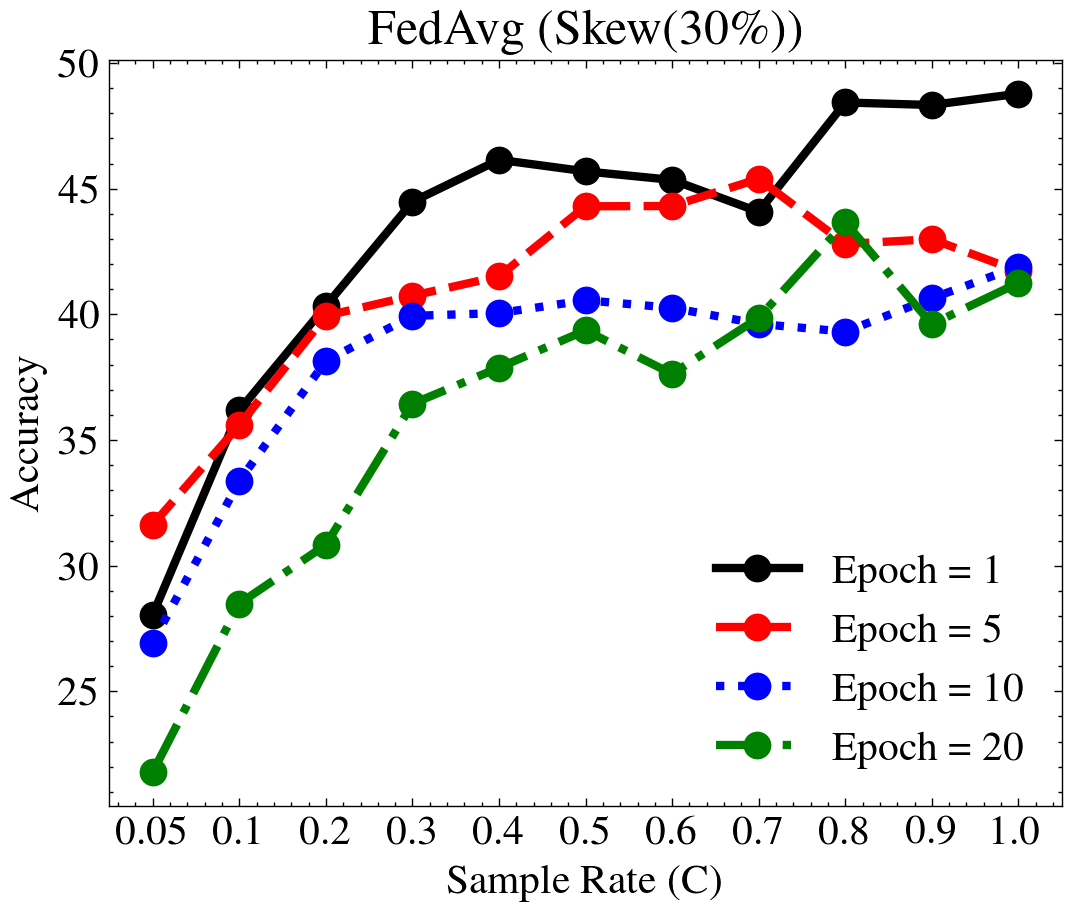}
         \caption{}
         \label{fig5:c}
     \end{subfigure}
      \begin{subfigure}{.24\textwidth}
         \centering
         \includegraphics[width=.85\linewidth]{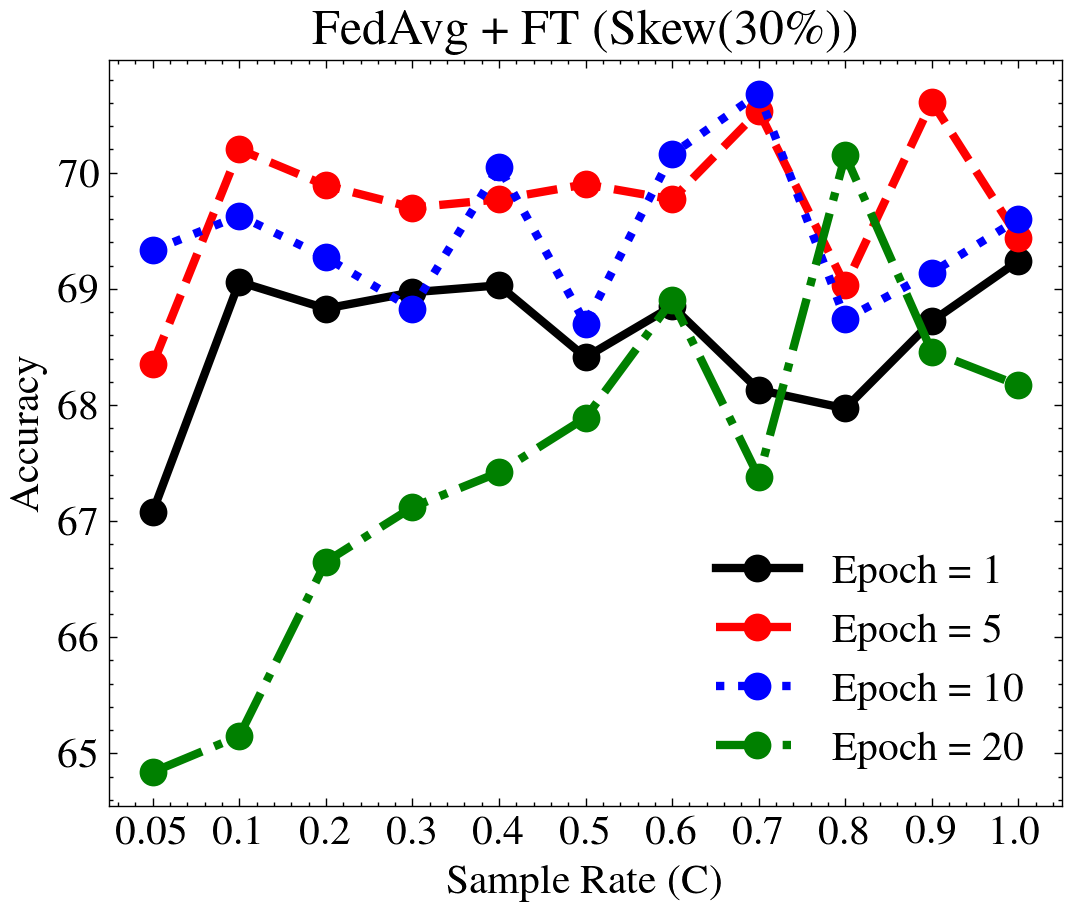}
         \caption{}
         \label{fig5:d}
     \end{subfigure}

    \caption{These figures illustrate the impact of sample rate ($C$) and local epochs on the performance: (a), (b) and (c), (d) show the performance of FedAvg and FedAvg + FT for Non-IID Label Dir($0.1$) and Non-IID Label Skew($30$\%), respectively.}
    \label{fig5}
    \vspace{-0.4cm}
\end{figure*}

Figures \ref{fig5:a}, \ref{fig5:b} and \ref{fig5:c},~\ref{fig5:d} demonstrate the impact of sample rate and local epochs on the performance for label Dir($0.1$) and label skew ($30$\%) types of statistical heterogeneity, respectively. Our observations show that \textbf{increasing the sample rate (i.e., averaging more models) can effectively \textcolor{red}{mitigate} the \textcolor{red}{negative} impact of statistical heterogeneity on performance}\footnote{\textcolor{red}{Model averaging is a standard component of many FL algorithms.}}. \textcolor{red}{Additionally, it is essential to consider that averaging a higher number of models (i.e., a high sample rate) reduces the approximation error associated with the averaged models across all clients\footnote{\textcolor{red}{It is also worth mentioning that generally high sample rates are not favorable, as they would increase the communication cost of the FL algorithms.}}.} We find that sample rates of $C >= 0.4$ can significantly reduce the effect of heterogeneity on the performance, while very small sample rates of $C < 0.1$ result in poor performance. Based on these findings, we suggest using a sample rate in the range of $0.1 <= C < 0.4$ for experimental design to accurately evaluate an algorithm's success in \textcolor{red}{the} presence of data heterogeneity.

% \begin{tcolorbox}[colback=white!5!white,colframe=black!75!black]
%   \textbf{Finding 4} (Effect of sample rate). We find that higher sample rate can indeed improve the results for all number of local epochs. Moreover, this improvement is more for lower than 5 local epochs. Also, hyper-parameter tuning becomes more difficult for higher than 0.5 number of local epochs.\footnote{We observe similar exists for other level of heterogeneity.}
% \end{tcolorbox}

\section{Summary and Recommendations}\label{recommendation}
In this section we identify a series of best practices and recommendations \textcolor{red}{for} designing a well-incentivized FL experimental setting base on our findings and insights in Section~\ref{study}.

\textbf{Level of statistical heterogeneity and local epochs.} The level of statistical heterogeneity and \textcolor{red}{the} number of local epochs determine the incentives for global and personalized FL. Generally, when the level of statistical heterogeneity is low, global FL is more incentivized than personalized FL, and vice versa (Figures~\ref{fig2:a} and~\ref{fig3:a}). Additionally, increasing the number of local epochs can make personalized FL more incentivized (Figures~\ref{app:fig-incentive-dir} and~\ref{app:fig-incentive-skew}). We have identified the well-incentivized settings for each FL approach in Table~\ref{tab:incentivized-settings}.

\begin{table}[t]
\caption{Well-incentivized settings for personalized and global FL approaches.}
\label{tab:incentivized-settings}
\centering
\resizebox{.6\linewidth}{!}{
\begin{tabular}{l|cc|cc|}
            \toprule
\multirow{2}{*}{Local Epoch} & \multicolumn{2}{c|}{Non-IID Label Dir} & \multicolumn{2}{c|}{Non-IID Label Skew}\\
          \cmidrule{2-3} \cmidrule{4-5}
 & pFL & gFL & pFL & gFL \\
            \midrule
$E=1$  & $\alpha < 0.3$ & $\alpha > 0.3$ & $p < 0.8$ & $p > 0.8$\\
            \midrule   
$E=5$  & $\alpha < 0.3$ & $\alpha > 0.3$ & $p < 0.8$ & $p > 0.8$\\
            \midrule
$E=10$  & $\alpha < 0.5$ & $\alpha > 0.5$ & $p < 0.8$ & $p > 0.8$\\
            \midrule
$E=20$  & $\alpha < 0.5$ & $\alpha > 0.5$ & $p < 0.9$ & $p > 0.9$\\
            \midrule
\end{tabular}
}
\end{table}

\textbf{Type of statistical heterogeneity.} We observe that the nature of label skew type of statistical heterogeneity favors personalized FL over a wider range of heterogeneity levels compared to label Dir (Figures~\ref{fig2:a},~\ref{fig3:a},~\ref{app:fig-incentive-dir} and~\ref{app:fig-incentive-skew}). Additionally, we observe that the impact of local epochs on performance is more pronounced for label Dir type of statistical heterogeneity compared to label skew (Figures~\ref{fig2:b},~\ref{fig2:c},~\ref{fig3:b} and~\ref{fig3:c}). To provide a comprehensive perspective on an algorithm's success in the presence of statistical heterogeneity, we recommend researchers conduct experiments with both types of statistical heterogeneity.

\textbf{Sample rate.} This variable plays a crucial role in evaluating an algorithm's performance under statistical heterogeneity. 
%Choosing a high sample rate ($C > 0.4$) can mask the effect of statistical heterogeneity and misrepresent an algorithm's ability to handle it, while a low sample rate ($C < 0.1$) can hinder convergence due to a lack of enough models to average (Figure~\ref{fig5}). 
\textcolor{red}{Choosing a high sample rate ($C > 0.4$) can mask the effect of statistical heterogeneity, thus misrepresenting an algorithm's true ability to handle it. Additionally, it can lead to inaccurate representations of an algorithm's capability to handle the stochasticity resulting from random device selection and the inherent errors caused by model averaging approximation. On the other hand, a low sample rate ($C < 0.1$) may hinder convergence due to insufficient models for averaging and high errors caused by model averaging approximation (Figure~\ref{fig5}).} To avoid these pitfalls, we recommend researchers use a sample rate of $0.1 \leq C \leq 0.4$ for their experiments.
% choosing a high sample rate ($C > 0.4$) can mask the effect of statistical heterogeneity and misrepresent an algorithm's true ability to handle it on one hand and lead to inaccurate representation of an algorithm's ability to handle the stochasticity resulting from random device selection and the inherent errors caused by model averaging approximation on the other hand. while a low sample rate ($C < 0.1$) can hinder convergence due to a lack of enough models to average and high errors caused by model averaging approximation. 

\textbf{Summary.} To ensure consistency, comparability, and meaningful results in FL experiments, we have compiled a set of recommended settings in Table~\ref{tab:recommended-settings}. We encourage researchers to adopt these settings, as well as the evaluation metrics outlined in Section~\ref{def:gfl}, for their experiments. This will facilitate more consistent and fair comparisons with SOTA algorithms, and eliminate concerns about evaluation failures and the impact of various experimental settings.

\begin{table}[ht]
\caption{Recommended settings for pFL and gFL approaches.}
\label{tab:recommended-settings}
\centering
\resizebox{1.0\linewidth}{!}{
\begin{tabular}{c|ccccc}
            \toprule
Approach & Type of Heterogeneity & Level of Heterogeneity & Local Epoch & Number of Clients & Sample Rate\\
            \midrule
\multirow{2}{*}{pFL} &  Label Dir & $\alpha \in [0.01, 0.3]$ & $\{1, 5, 10, 20\}$ & $\{20, 100\}$ & $\{0.1, 0.2, 0.3, 0.4\}$ \\
            \cmidrule{2-6}
  & Label Skew & $p \in \{2, 3, 4, 5\}$ & $\{1, 5, 10, 20\}$ & $\{20, 100, 500\}$ & $\{0.1, 0.2, 0.3, 0.4\}$ \\
            \midrule
\multirow{2}{*}{gFL} &  Label Dir & $\alpha \in (0.3, 1]$ & $\{1, 5, 10, 20\}$ & $\{20, 100\}$ & $\{0.1, 0.2, 0.3, 0.4\}$ \\
            \cmidrule{2-6}
  & Label Skew & $p \in \{8, 9\}$ & $\{1, 5, 10, 20\}$ & $\{20, 100, 500\}$ & $\{0.1, 0.2, 0.3, 0.4\}$ \\
  \bottomrule
\end{tabular}
}
\vspace{-0.5cm}
\end{table}

\begin{comment}
\begin{table}[ht]
\caption{Recommended settings for pFL and gFL approaches.}
\label{tab:recommended-settings}
\centering
\resizebox{0.9\linewidth}{!}{
\begin{tabular}{c|ccccc}
            \toprule
Approach & Type of Heterogeneity & Level of Heterogeneity & Local Epoch & Number of Clients & Sample Rate\\
            \midrule
\multirow{2}{*}{pFL} &  Label Dir & $\alpha \in [0.01, 0.3]$ & $E \in \{1, 5, 10, 20\}$ & $N \in \{20, 100\}$ & $S \in \{0.1, 0.2, 0.3, 0.4\}$ \\
            \cmidrule{2-6}
  & Label Skew & $p \in \{2, 3, 4, 5\}$ & $E \in \{1, 5, 10, 20\}$ & $N \in \{20, 100, 500\}$ & $S \in \{0.1, 0.2, 0.3, 0.4\}$ \\
            \midrule
\multirow{2}{*}{gFL} &  Label Dir & $\alpha \in [0.3, 1]$ & $E \in \{1, 5, 10, 20\}$ & $N \in \{20, 100\}$ & $S \in \{0.1, 0.2, 0.3, 0.4\}$ \\
            \cmidrule{2-6}
  & Label Skew & $p \in \{8, 9\}$ & $E \in \{1, 5, 10, 20\}$ & $N \in \{20, 100, 500\}$ & $S \in \{0.1, 0.2, 0.3, 0.4\}$ \\
  \bottomrule
\end{tabular}
}
\end{table}
\end{comment}
\section{FedZoo-Bench}\label{fedzoo}
We introduce FedZoo-Bench, an open-source library based on PyTorch that facilitates experimentation in federated learning by providing researchers with a comprehensive set of standardized and customizable features such as training, Non-IID data partitioning, fine-tuning, performance evaluation, fairness assessment, and generalization to newcomers, for both global and personalized FL approaches. Additionally, it comes pre-equipped with a set of models and datasets, \textcolor{red}{and} pre-implemented 22 different SOTA methods, allowing researchers to quickly test their ideas and hypotheses. FedZoo-Bench is a powerful tool that empowers researchers to explore new frontiers in federated learning and \textcolor{red}{enables} fair, consistent, and reproducible research in federated learning. We have provided more details on the implemented algorithms and available features in Appendix Section~\ref{sec:app-fedzoo}.

\subsection{Comparison of SOTA methods}\label{sec:comparison}
In this section, we present a comprehensive experimental comparison of several SOTA FL methods using FedZoo-Bench. We evaluate their performance, fairness, and generalization to new clients \textcolor{red}{in} a consistent experimental setting. Our experiments aim to provide a better understanding of the current progress in FL research.

\textbf{Training setting.} Following our recommended settings indicated in Table~\ref{tab:recommended-settings}, we choose two different training settings presented in Table~\ref{tab:training-setting} to conduct the experimental comparison of the baselines for each FL approach. We run each of the baselines 3 times and report the average and standard deviation of the results. 
%For more details of the hyperparameters used for each baseline, see Appendix Section~\ref{sec:app-hyperparameters}.

\begin{table}[hb]
\caption{Training settings}
\centering
\resizebox{0.5\textwidth}{!}{
\begin{tabular}{c|ccccccc}
\toprule
Setting           & Dataset & Architecture & clients & sample rate & local epochs & partitioning & communication rounds  \\
\midrule
gFL \#1 & CIFAR-10 & LeNet-5 & 100 & 0.1 & 5 & Non-IID Label Skew(80\%) & 100\\
\midrule
gFL \#2 & CIFAR-100 & ResNet-9 & 20 & 0.2 & 10 & Non-IID Label Dir(0.5) & 100 \\
\midrule
pFL \#1 & CIFAR-10 & LeNet-5 & 100 & 0.1 & 10 & Non-IID Label Skew(30\%) & 100\\
\midrule
pFL \#2 & CIFAR-100 & ResNet-9 & 20 & 0.2 & 10 & Non-IID Label Dir(0.1) & 100 \\
\bottomrule
\end{tabular}
}
\label{tab:training-setting}
\end{table}

\begin{table}[ht]
\caption{Performance and fairness comparison for personalized FL baselines.}
\begin{center}
\begin{subtable}[c]{.8\linewidth}
\caption{Performance Comparison}
\label{tab:pfl-acc}
\centering
\resizebox{0.8\linewidth}{!}{
\begin{tabular}{l|c|c}
            \toprule
Algorithm & Setting (pFL \#1) & Setting (pFL \#2)\\
            \midrule
FedAvg + FT~\cite{jiang2019improving}  & $69.26\pm0.42$ & $49.03\pm0.40$  \\
            \midrule         
LG-FedAvg~\cite{liang2020think}  & $54.03\pm2.16$ & $25.30\pm0.50$\\
            \midrule  
PerFedAvg~\cite{fallah2020personalized}  & $76.03\pm0.31$ & $2.29\pm0.07$ \\
            \midrule
IFCA~\cite{ghosh2020efficient}  & $64.84\pm3.41$ & $46.73\pm1.82$ \\
            \midrule
Ditto~\cite{li2021ditto}  & $70.97\pm1.27$ & $48.16\pm3.25$ \\
            \midrule
FedPer~\cite{arivazhagan2019federated}  & $64.64\pm0.45$ & $42.87\pm1.66$ \\
            \midrule
FedRep~\cite{collins2021exploiting}  & $54.99\pm 3.16$ & $29.39\pm0.31$ \\
            \midrule
APFL~\cite{deng2020adaptive}  & $68.91\pm0.59$ & $55.18\pm0.65$ \\
            \midrule
pFedMe~\cite{t2020personalized}  & $10.00\pm0.98$ & $2.10\pm0.30$ \\
            \midrule
CFL~\cite{sattler2020clustered}  & $16.83\pm1.6$ & $1.52\pm0.17$ \\
            \midrule
SubFedAvg~\cite{vahidian2021personalized} & $69.95\pm1.34$ & $49.83\pm1.09$ \\
            \midrule
PACFL~\cite{vahidian2022efficient}  & $67.78\pm0.11$ & $50.11\pm1.10$ \\
            \bottomrule
\end{tabular}
}
\vspace{0.2cm}
\end{subtable}
\end{center}
\begin{center}
\begin{subtable}[c]{.8\linewidth}
\caption{Fairness Comparison}
\label{tab:pfl-fairness}
\centering
\resizebox{0.8\linewidth}{!}{
\begin{tabular}{l|c|c}
            \toprule
Algorithm & Setting (pFL \#1) & Setting (pFL \#2) \\
            \midrule
FedAvg + FT~\cite{jiang2019improving}  & $8.05\pm0.32$ & $5.40\pm0.77$  \\
            \midrule       
LG-FedAvg~\cite{liang2020think}  & $12.79\pm0.64$ & $5.11\pm0.38$\\
            \midrule  
PerFedAvg~\cite{fallah2020personalized}  & $7.32\pm0.27$ & $--$ \\
            \midrule 
IFCA~\cite{ghosh2020efficient}  & $10.56\pm2.75$ & $5.03\pm0.07$ \\
            \midrule  
Ditto~\cite{li2021ditto}  & $7.50\pm0.37$ & $4.10\pm0.73$ \\
            \midrule  
FedPer~\cite{arivazhagan2019federated}  & $7.64\pm0.59$ & $6.19\pm0.59$ \\
            \midrule  
FedRep~\cite{collins2021exploiting}  & $9.18\pm0.50$ & $5.19\pm0.73$ \\
            \midrule  
APFL~\cite{deng2020adaptive}  & $8.22\pm0.96$ & $4.60\pm1.15$ \\
            \midrule
pFedMe~\cite{t2020personalized}  & $--$ & $--$ \\
            \midrule
SubFedAvg~\cite{vahidian2021personalized} & $7.44\pm0.46$ & $4.66\pm0.56$ \\
            \midrule
CFL~\cite{sattler2020clustered} & $--$ & $--$ \\
            \midrule
PACFL~\cite{vahidian2022efficient} & $8.82\pm0.71$ & $4.68\pm0.42$ \\
            \bottomrule
\end{tabular}
}
\end{subtable}
\end{center}
\end{table}

\textbf{Performance comparison.} We use the evaluation metrics outlined in Section~\ref{def:gfl} to compare the performance results. \vspace{-0.1cm}
\begin{itemize}
    \item \textbf{Global FL.} Table~\ref{tab:gfl-acc} shows the performance results of 6 different global FL methods. As it is noticeable Scaffold and FedProx \textcolor{red}{have} given the best results in settings (gfl \#1) and (gfl \#2), respectively and FedDF has given the worst results in both settings. FedAvg which is the simplest method has appeared to be competitive in both settings and even better than 4 other algorithms in setting (gfl \#2).
    \item \textbf{Personalized FL.} In Table~\ref{tab:pfl-acc}, we present the performance results of 12 different personalized FL methods. Similar to global FL methods (Table~\ref{tab:gfl-acc}), we observe that each method performs differently in different settings. No single method consistently achieves the best results across all settings. For example, PerFedAvg performs well in setting (pfl \#1), but poorly in setting (pfl \#2). Additionally, CFL and pFedMe perform poorly in both settings. On the other hand, FedAvg + FT, the simplest baseline, performs fairly well in both settings and is competitive or even superior to several other methods.
\end{itemize} 

\textbf{Fairness comparison.} Fairness is another important aspect \textcolor{red}{of the} personalized FL approach. We use the fairness metric mentioned in~\cite{li2021ditto, mohri2019agnostic} which is the standard deviation of final local test accuracies. Table~\ref{tab:pfl-fairness} shows the fairness comparison of the methods. SubFedAvg and Ditto have achieved the best fairness results in (pfl \#1) and (pfl \#2) settings, respectively. FedAvg + FT also demonstrated competitive fairness results in both settings. For algorithms having poor results we did not report the fairness results as they would not be meaningful.

\textbf{Generalization to newcomers.} To evaluate the generalization capabilities of personalized FL methods to newcomers, we reserve 20\% of the clients as newcomers and train the FL models using the remaining 80\% clients. While the adaptation process for many methods is not explicitly clear, we follow the same procedure as in~\cite{marfoq2021federated, vahidian2022efficient} and allow the newcomers to receive the trained FL model and perform local fine-tuning. For methods like PACFL that have a different adaptation strategy, we follow their original approach. Table~\ref{tab:pfl-newcommers} shows the results of this evaluation.

\textbf{Discussion.} The experimental comparison between several SOTA methods for each FL approach outlined in this section highlights the progress that has been made in FL research. While we can see that many methods have improved compared to the simple FedAvg and FedAvg + FT baselines for global and personalized FL approaches, respectively, there are also some limitations that are worth noting:
\begin{itemize}
    \item There is no method that consistently performs the best across all experimental settings. Furthermore, for the personalized FL approach, a method may achieve good fairness results but lack generalization to newcomers. Thus, evaluating FL methods from different perspectives and developing algorithms that can provide a better trade-off is crucial.
    \item Despite the existence of numerous works for each FL approach, the performance of the simple methods of FedAvg and FedAvg + FT are still competitive or even better compared to several methods. Thus, there is a need for new methods that can achieve consistent improvements across different types of statistical heterogeneity.
    \item Fairness and generalization to newcomers are two important aspects of the personalized FL approach that are often overlooked in the literature and \textcolor{red}{are} only focused on performance improvement. Therefore, it is crucial to consider these aspects in addition to performance improvement when designing new personalized FL methods.
\end{itemize}

\begin{minipage}[c][][b]{0.5\textwidth}
\centering
\captionof{table}{gFL accuracy comparison}
\label{tab:gfl-acc}
\resizebox{0.6\linewidth}{!}{
\begin{tabular}{l|c|c}
            \toprule
Algorithm & Setting (gFL \#1) & Setting (gFL \#2)\\
            \midrule
FedAvg~\cite{mcmahan2017communication}  & $44.89\pm0.20$ & $56.47\pm0.57$\\
            \midrule
FedProx~\cite{li2020federated}  & $46.01\pm0.46$ & $56.85\pm0.36$\\
            \midrule
FedNova~\cite{wang2020tackling}  & $44.59\pm0.60$ & $53.20\pm0.33$\\
            \midrule
Scaffold~\cite{karimireddy2020scaffold}  & $56.85\pm1.06$ & $51.71\pm0.65$\\
            \midrule
FedDF~\cite{lin2020ensemble} & $27.43\pm 2.32$ & $30.24\pm0.26$ \\
            \midrule
MOON~\cite{li2021model} & $45.60\pm0.31$ & $50.23\pm0.55$ \\
            \bottomrule
\end{tabular}
}
\vspace{0.2cm}
\end{minipage}
\begin{minipage}[c][][b]{0.5\textwidth}
\centering
\captionof{table}{pFL generalization to newcomers}
\label{tab:pfl-newcommers}
\resizebox{0.6\linewidth}{!}{
\begin{tabular}{l|c|c}
            \toprule
Algorithm & Setting (pFL \#1) & Setting (pFL \#2)\\
            \midrule
FedAvg + FT~\cite{jiang2019improving} & $64.19\pm4.64$ & $37.14\pm0.43$ \\
            \midrule
LG-FedAvg~\cite{liang2020think}   & $40.39\pm17.98$ & $21.22\pm2.56$ \\
            \midrule  
PerFedAvg~\cite{fallah2020personalized}   & $74.97\pm1.10$ & $2.22\pm0.20$ \\
            \midrule  
IFCA~\cite{ghosh2020efficient}   & $62.64\pm1.03$ & $14.84\pm1.86$ \\
            \midrule
Ditto~\cite{li2021ditto}   & $62.55\pm3.10$ & $38.96\pm0.26$ \\
            \midrule
FedPer~\cite{arivazhagan2019federated}   & $65.3\pm2.41$ & $35.66\pm1.61$ \\
            \midrule
FedRep~\cite{collins2021exploiting}   & $64.50\pm0.62$ & $23.85\pm1.49$ \\
            \midrule
APFL~\cite{deng2020adaptive}   & $66.38\pm1.25$ & $39.52\pm1.11$ \\
            \midrule
SubFedAvg~\cite{vahidian2021personalized} & $63.54\pm1.42$ & $30.81\pm1.28$ \\
            \midrule
PACFL~\cite{vahidian2022efficient} & $68.54\pm1.33$ & $36.50\pm1.42$ \\
            \bottomrule
\end{tabular}
}
\end{minipage}

\section{Conclusion and Future Works}\label{sec:conclusion}
In this paper, we present a thorough examination of key variables that influence the success of FL experiments. Firstly, we provide new insights and analysis of the FL-specific variables in relation to each other and performance results by running several experiments. We then use our analysis to identify recommendations and best practices for a meaningful and well-incentivized FL experimental design. We have also developed FedZoo-Bench, an open-source library based on PyTorch that provides a comprehensive set of standardized and customizable features, different evaluation metrics, and implementation of 22 SOTA methods. FedZoo-Bench facilitates a more consistent and reproducible FL research. Lastly, we conduct a comprehensive evaluation of several SOTA methods in terms of performance, fairness, and generalization to newcomers using FedZoo-Bench. We hope that our work will help the FL community to better understand the state of progress in the field and encourage a more comparable and consistent FL research.

In future work, we plan to expand our study to other domains such as natural language processing and graph neural networks to understand how FL experimental settings behave in those areas \textcolor{red}{and to assess its versatility and applicability across different problem domains.} Additionally, we will continue to improve our benchmark by implementing more algorithms and adding new features. \textcolor{red}{We also have plans to establish an open leaderboard using FedZoo-Bench, enabling systematic evaluations of FL methods across a wide variety of datasets and settings.} Based on the comparison results presented in Section~\ref{sec:comparison}, we also believe that \textcolor{red}{the} development of new algorithms for both global and personalized FL approaches to achieve even greater improvement and more consistent results across different experimental settings would be an exciting future avenue. Also, more studies on evaluation metrics and \textcolor{red}{the} development of new metrics that can better assess different aspects of an FL algorithm would be a valuable future work.

%In future work, we plan to expand our study to other domains such as natural language processing and graph neural networks to understand how FL experimental variables behave for other ML tasks. Additionally, we will continue to improve our benchmark by implementing more algorithms and adding new features. Based on the comparison results presented in Section~\ref{sec:comparison}, development of new algorithms for both global and personalized FL approaches that can achieve even greater improvement and consistent results across different experimental settings would be an exciting avenue for future research. Also, further research and studies on evaluation metrics and the development of new metrics that capture different aspects of federated learning would be valuable contributions to the field.
%\input{algorithms}

%\appendix
\section*{APPENDIX}
\appendices
\textbf{Organization.} We organize the supplementary materials as follow: 
\begin{itemize}
    \item In Section~\ref{app:sec-additional}, we present addition experimental results to complete our analysis in Section~\ref{study} of the main paper.
    \item In Section~\ref{app:recommendation}, we discuss an experimental checklist to facilitate an easier comparison of FL methods.
    \item In Section~\ref{sec:app-fedzoo}, we provide more details about the available algorithms, datasets, architectures and data partitionings in FedZoo-Bench.
    %\item In Section~\ref{sec:app-hyperparameters}, we brings the details of the hyperparameters we used for each method in our experiments of Section~\ref{sec:comparison}. 
\end{itemize}
\section{Additional Results}\label{app:sec-additional}
%In this section we present additional experimental results for completeness of our study.
\subsection{Globalization and Personalization Incentives} \label{appendix:incentives}
\begin{figure*}[htbp]
    \centering
    \begin{subfigure}{.24\textwidth}
         \centering
         \includegraphics[width=.99\linewidth]{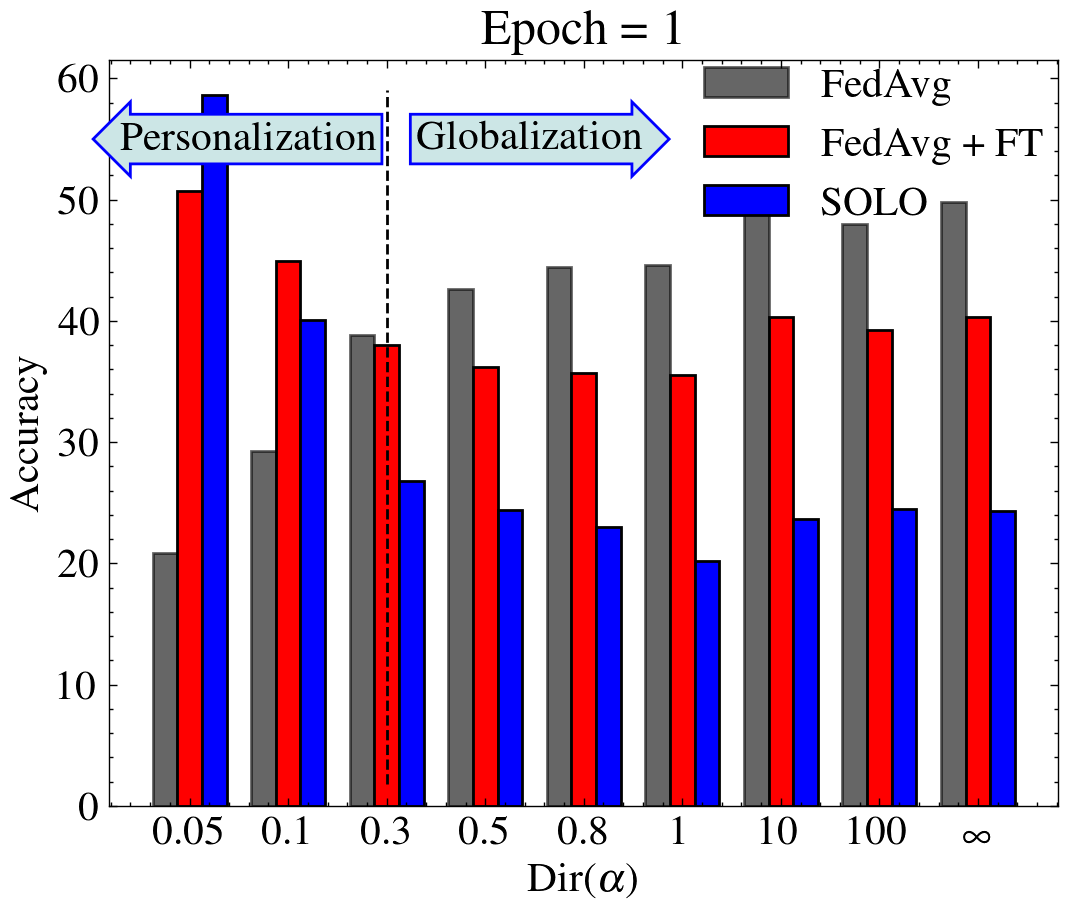}
         %\caption{}
         %\label{figa:a}
     \end{subfigure}
     \begin{subfigure}{.24\textwidth}
         \centering
         \includegraphics[width=.99\linewidth]{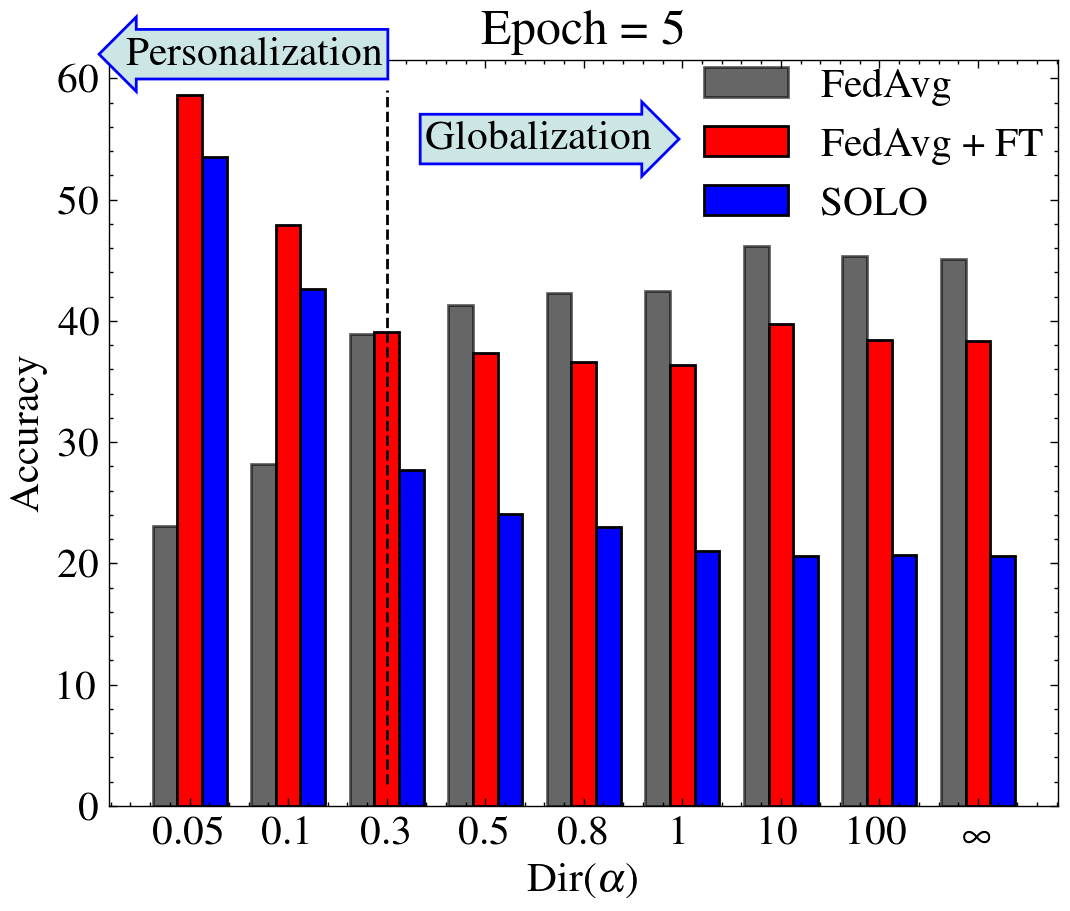}
         %\caption{}
         %\label{figa:b}
     \end{subfigure}
     \begin{subfigure}{.24\textwidth}
         \centering
         \includegraphics[width=.99\linewidth]{figures/v2/wtp_dir_e10.png}
         %\caption{}
         %\label{figa:c}
     \end{subfigure}
      \begin{subfigure}{.24\textwidth}
         \centering
         \includegraphics[width=.99\linewidth]{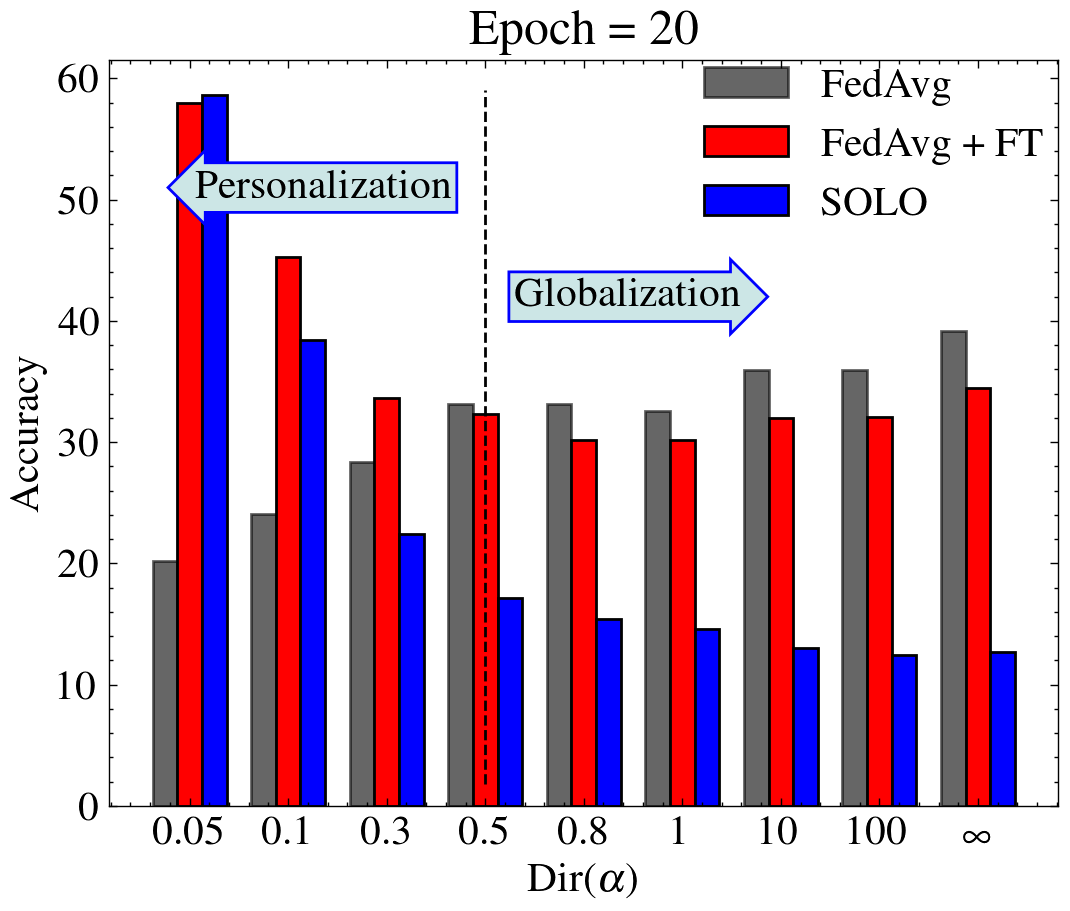}
         %\caption{}
         %\label{figa:d}
     \end{subfigure}

    \caption{These figures show globalization and personalization incentives at different level of heterogeneity and local epochs for Non-IID Label Dir partitioning. The approximate boundary shifts from $0.3$ to $0.5$ with the increase of local epochs.}
    \label{app:fig-incentive-dir}
    \vspace{-0.5cm}
\end{figure*}

\begin{figure*}[htbp]
    \centering
    \begin{subfigure}{.24\textwidth}
         \centering
         \includegraphics[width=.99\linewidth]{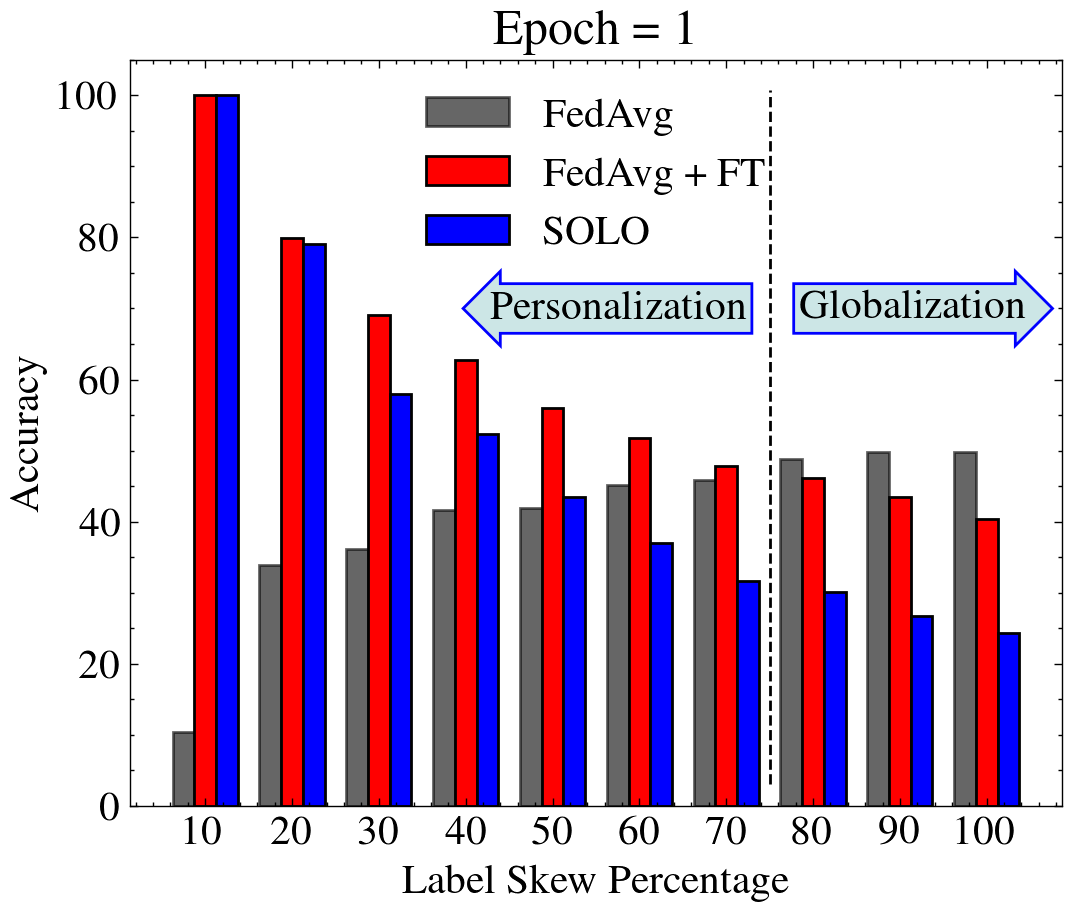}
         %\caption{}
         %\label{figb:a}
     \end{subfigure}
     \begin{subfigure}{.24\textwidth}
         \centering
         \includegraphics[width=.99\linewidth]{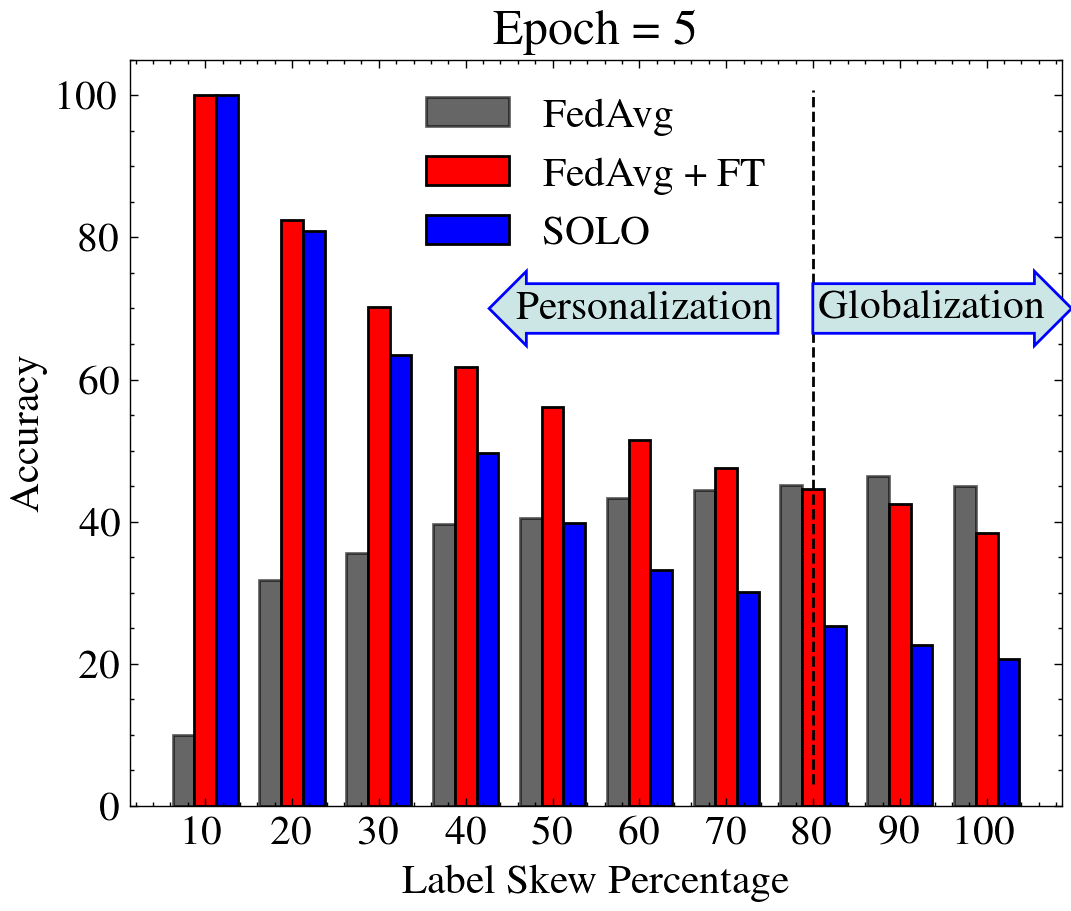}
         %\caption{}
        % \label{figb:b}
     \end{subfigure}
     \begin{subfigure}{.24\textwidth}
         \centering
         \includegraphics[width=.99\linewidth]{figures/v2/wtp_ls_e10.png}
         %\caption{}
         %\label{figb:c}
     \end{subfigure}
      \begin{subfigure}{.24\textwidth}
         \centering
         \includegraphics[width=.99\linewidth]{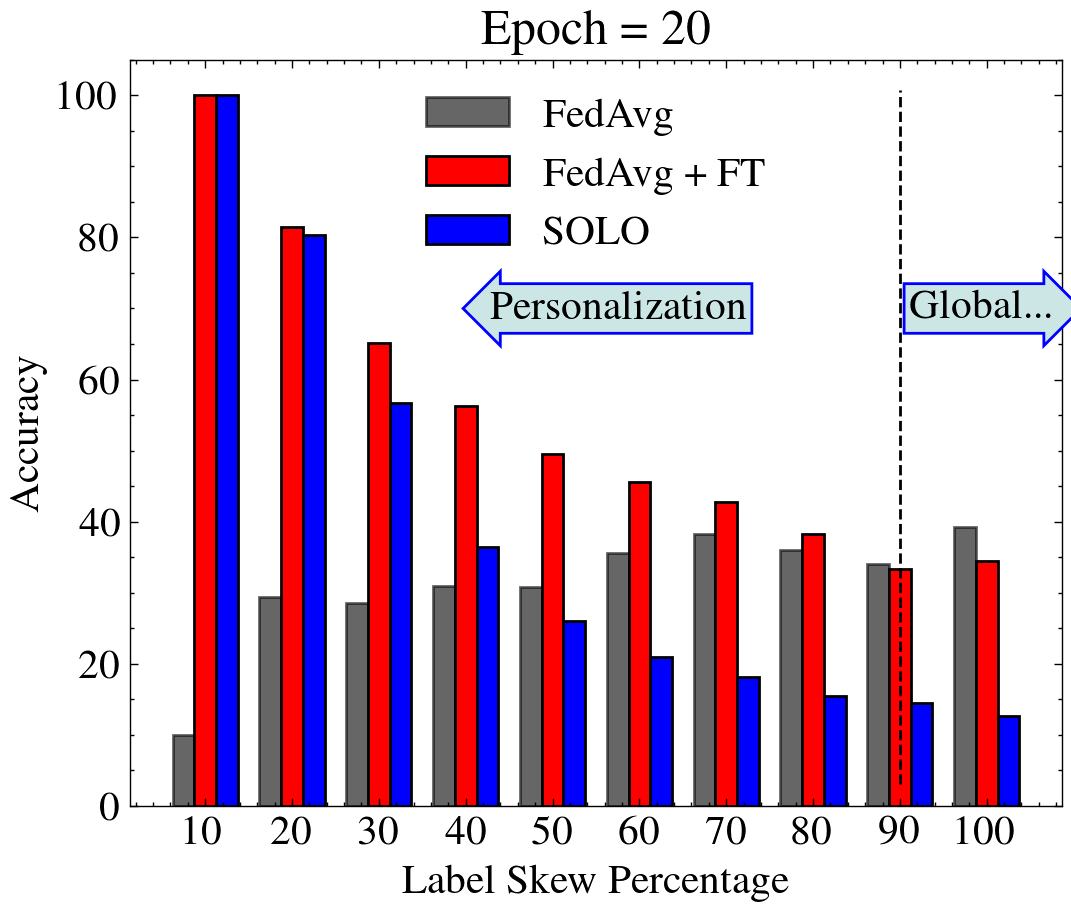}
         %\caption{}
         %\label{figb:d}
     \end{subfigure}

    \caption{These figures show globalization and personalization incentives at different level of heterogeneity and local epochs for Non-IID Label Skew partitioning. The approximate boundary shifts from $80\%$ to $90\%$ with the increase of local epochs.}
    \label{app:fig-incentive-skew}
\end{figure*}

The additional results in this part complements the results discussed in Section~\ref{sec:heterogeneity-localepoch}. Comparing Figures~\ref{app:fig-incentive-dir} and~\ref{app:fig-incentive-skew} further corroborates our finding mentioned in Section~\ref{sec:heterogeneity-localepoch} that Non-IID Label Skew partitioning has higher incentive for personalization compared to the other type of heterogeneity. Moreover, increase of local epochs has incentivized personalization more for both types of heterogeneity.

\begin{comment}
\subsection{Experimental Settings}

\begin{table}[h]
\caption{Hyper-parameters used for each figure}
\centering
\resizebox{0.8\textwidth}{!}{
\begin{tabular}{c|cccccccccccc}
\toprule
Figure           & Dataset & Architecture & clients & sample rate & epochs & communication rounds & optimizer & learning rate & momentum  & partitioning & mics \\
\midrule
Figure 1 & XX & XX & XX & XX & XX & XX & XX & XX & XX & XX & XX \\
\midrule
Figure 1 & XX & XX & XX & XX & XX & XX & XX & XX & XX & XX & XX \\
\midrule
Figure 1 & XX & XX & XX & XX & XX & XX & XX & XX & XX & XX & XX \\
\midrule
Figure 1 & XX & XX & XX & XX & XX & XX & XX & XX & XX & XX & XX \\
\midrule

\bottomrule
\end{tabular}
}
\label{tab:hyperparameters-figs-tabs}
\end{table}
\end{comment}

\section{Experimental Checklist} \label{app:recommendation}
To facilitate an easier comparison between FL methods in future studies, we recommend the following checklist:
\begin{itemize}
    \item Make sure that the used experimental setting is meaningful and well-incentived for the considered FL approach. 
    \item State the exact setting including local epochs, sample rate, number of clients, type of data partitioning, level of heterogeneity, communication rounds, dataset, architecture, evaluation metrics, any pre-processing on the dataset, any learning rate scheduling if used, and initialization.
    \item Report the average results over at least $3$ independent and different runs.
    \item Mention the hyperparameters used to obtain the results.
\end{itemize}

\section{FedZoo-Bench}\label{sec:app-fedzoo}
We introduced FedZoo-Bench in Section~\ref{fedzoo}. In this section we provide more details about the available features in FedZoo-Bench. For more information on FedZoo-Bench's implementation and use cases for different settings, refer to the project's documentation at~\url{https://github.com/MMorafah/FedZoo-Bench}.
\subsection{Available Baselines}
\begin{itemize}
    \item \textbf{Global FL} (7 algorithms)
    \begin{itemize}
        \item FedAvg~\cite{mcmahan2017communication}
        \item FedProx~\cite{li2020federated}
        \item FedNova~\cite{wang2020tackling}
        \item Scaffold~\cite{karimireddy2020scaffold}
        \item FedDF~\cite{lin2020ensemble}
        \item MOON~\cite{li2021model}
        \item FedBN~\cite{li2021fedbn}
    \end{itemize}
    \item \textbf{Personalized FL} (15 algorithms)
    \begin{itemize}
        \item FedAvg + FT~\cite{jiang2019improving}
        \item LG-FedAVg~\cite{liang2020think}
        \item PerFedAvg~\cite{fallah2020personalized}
        \item FedPer~\cite{arivazhagan2019federated}
        \item FedRep~\cite{collins2021exploiting}
        \item Ditto~\cite{li2021ditto}
        \item APFL~\cite{deng2020adaptive}
        \item IFCA~\cite{ghosh2020efficient}
        \item SubFedAvg~\cite{vahidian2021personalized}
        \item pFedMe~\cite{t2020personalized}
        \item CFL~\cite{sattler2020clustered}
        \item PACFL~\cite{vahidian2022efficient}
        \item MTL~\cite{smith2017federated}
        \item FedEM~\cite{marfoq2021federated}
        \item FedFOMO~\cite{zhang2020personalized}
    \end{itemize}
\end{itemize}
Additionally, FedZoo-Bench can be easily used for other variations of FedAvg~\cite{reddi2020adaptive} and different choice of optimizers.

\subsection{Available Datasets}
\begin{itemize}
    \item MNIST~\cite{mnist}
    \item CIFAR-10~\cite{cifar}
    \item CFIAR-100
    \item USPS~\cite{usps}
    \item SVHN~\cite{svhn}
    \item CelebA~\cite{celeba}
    \item FMNIST~\cite{fmnist}
    \item FEMNIST~\cite{femnist}
    \item Tiny-ImageNet~\cite{tiny}
    \item STL-10~\cite{stl-10}
\end{itemize}

\subsection{Available Architectures}
\begin{itemize}
    \item MLP as in FedAvg \cite{mcmahan2017communication}
    % should we give a specific architecture of mlp?
    \item LeNet-5 \cite{lenet}
    \item ResNet Family \cite{resnet}
    \item ResNet-50
    % I think resnet-50 is more like part of resnet family than resnet-9. Maybe we should put resnet-9 here?
    \item VGG Family \cite{vgg}
\end{itemize}

\subsection{Available Data Partitionings}
\begin{itemize}
    \item IID
    \item Non-IID Label Dir~\cite{hsu2019measuring}
    \item Non-IID Label Skew~\cite{li2021federated}
    \item Non-IID Random Shard~\cite{mcmahan2017communication}
    \item Non-IID Quantity Skew~\cite{li2021federated}
\end{itemize}

\newpage
\clearpage
%\section*{References}
%\bibliography{Mahdi-Fed}
\bibliography{Non_IID, extra}
%\bibliography{extra}
% \newpage
% \clearpage

\begin{IEEEbiography}
    [{\includegraphics[width=1in,height=5in,clip,keepaspectratio]{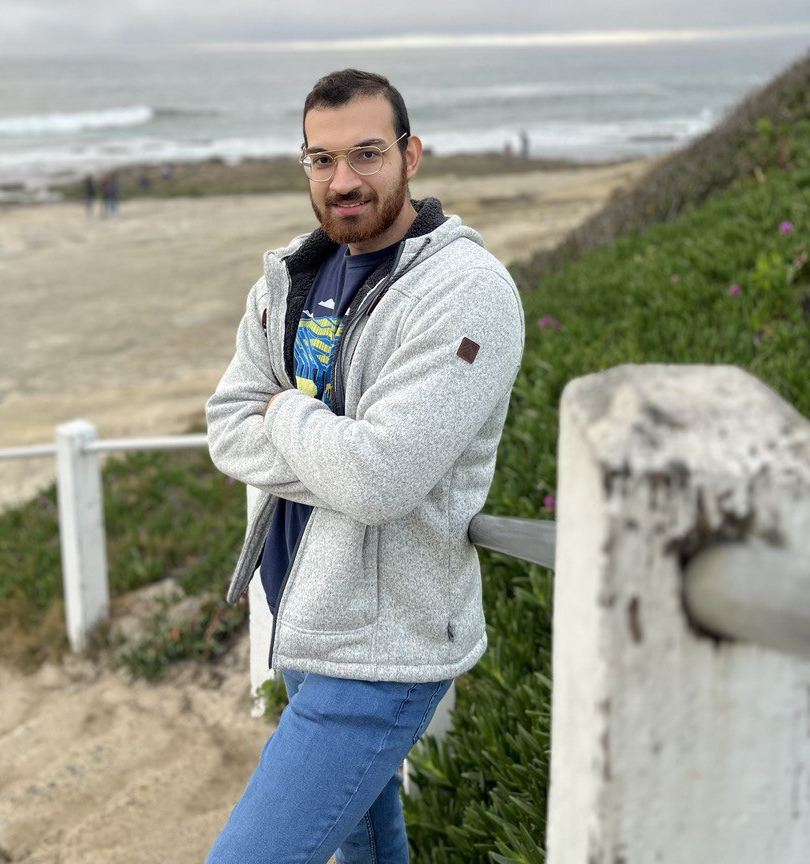}}]{Mahdi Morafah}
received the BS in Electrical Engineering from the Tehran Polytechnic University, Tehran, Iran, in 2019, and the MS degree in Electrical and Computer Engineering from the University of California, San Diego, in 2021. He is currently pursuing the PhD degree in Electrical and Computer Engineering at the University of California, San Diego. His research interests include Machine Learning and Deep Learning, Distributed Learning, Federated Learning, Continual Learning, and Optimization. 
\end{IEEEbiography}

\begin{IEEEbiography}
    [{\includegraphics[width=1in,height=5in,clip,keepaspectratio]{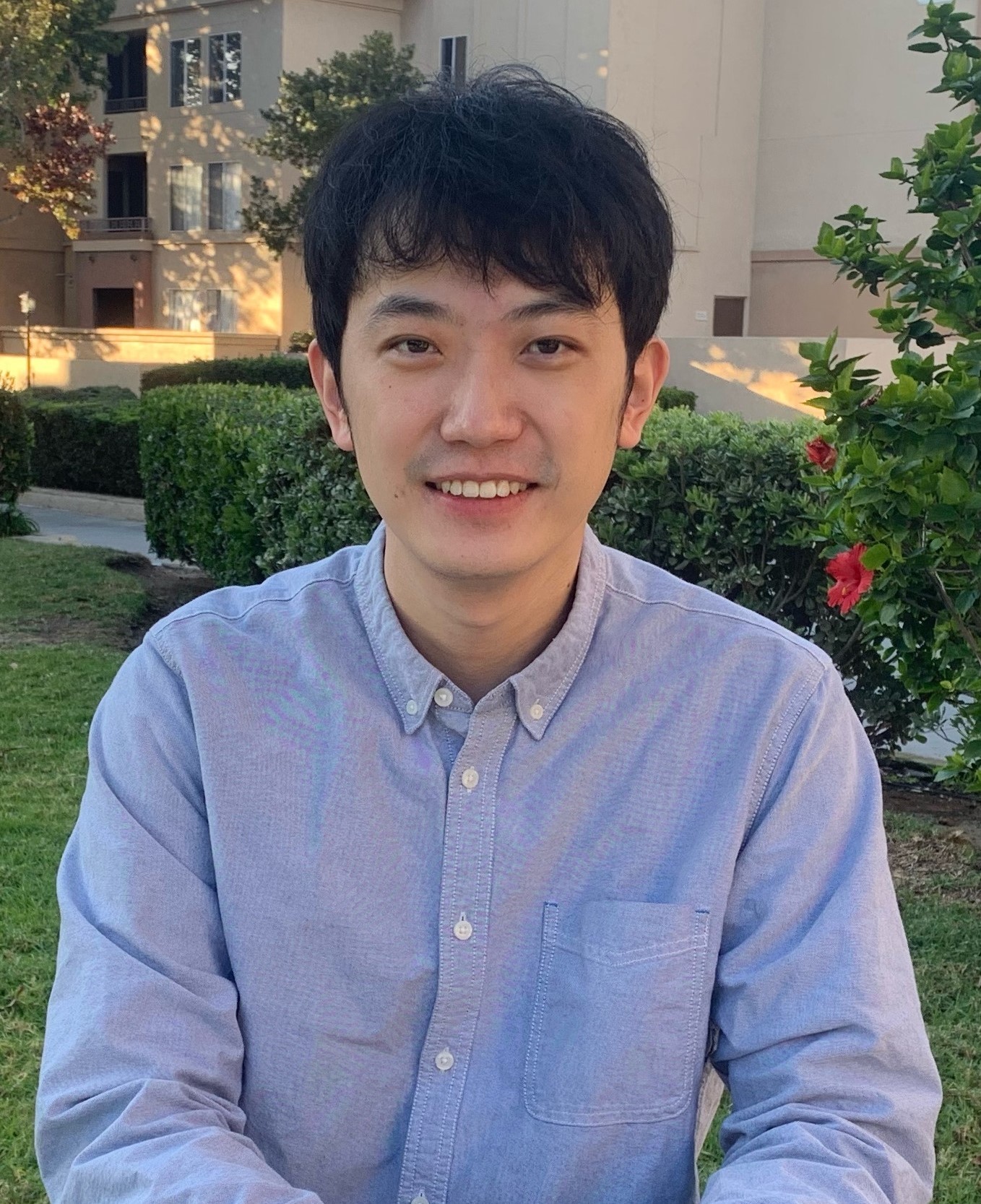}}]{Weijia Wang}
received the B.S. degree in Electrical Engineering from Zhejiang University, Hangzhou, China, in 2016 and the M.S. degree in Electrical and Computer Engineering from the University of California, San Diego, in 2018. He is currently pursuing the Ph.D. degree in Electrical and Computer Engineering at the University of California, San Diego. His research interest is machine learning and deep learning, including the compression and acceleration of deep convolutional neural networks, algorithms of meta-learning and federated learning, and explainable artificial intelligence.
\end{IEEEbiography}

\begin{IEEEbiography}
    [{\includegraphics[width=0.9in,height=4in,clip,keepaspectratio]{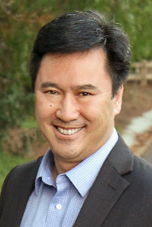}}]{Bill Lin}
received the BS, MS, and the PhD degrees in electrical engineering and computer sciences from the University of California, Berkeley in 1985, 1988, and 1991, respectively. He is a Professor in Electrical and Computer Engineering at the University of California, San Diego, where he is actively involved with the Center for Wireless Communications (CWC), the Center for Networked Systems (CNS), and the Qualcomm Institute in industry-sponsored research efforts. His research has led to over 200 journal and conference publications, including a number of Best Paper awards and nominations. He also holds 5 awarded patents. He has served as the General Chair and on the executive and technical program committee of many IEEE and ACM conferences, and he has served as an Associate or Guest Editor for several IEEE and ACM journals as well.
\end{IEEEbiography}

\end{document}